\documentclass[aps, prd, showpacs, floatfix, superscriptaddress, twocolumn, nofootinbib, preprintnumbers, longbibliography]{revtex4-2}
\usepackage{lipsum, multirow, microtype, amsmath, amssymb, newfloat, bm, color}
\usepackage{natbib}
\usepackage{graphicx}
\usepackage{comment}
\usepackage{amssymb}
\usepackage{hyperref}
\usepackage{hypcap, mathrsfs, placeins, etoolbox}
\usepackage{dcolumn}
\usepackage[dvipsnames]{xcolor}
\usepackage[T1]{fontenc}
\usepackage[utf8]{inputenc}
\usepackage[scaled = 0.92]{helvet}
\usepackage{inconsolata}
\usepackage{siunitx}
\usepackage{booktabs}
\usepackage{xurl}
\usepackage{lipsum, multirow, microtype, amsmath, amssymb, newfloat, bm, color}
\usepackage{lineno}
\usepackage{enumitem}
\setlist[itemize]{leftmargin = 9pt, labelsep = 0.31em, itemsep = 0.055em}
\graphicspath{{Figures/}} 

\hypersetup{
colorlinks = true,
citecolor = cyan,
linkcolor = magenta,
urlcolor = NavyBlue,
allbordercolors = {0 0 0},
pdfborderstyle = {/S/U/W 1}
}
\newcommand{\IIIR}{Department of Computer Science \& Engineering, Indian Institute of Information Technology Raichur, Karnataka - 584135, India.\vspace*{0.125cm}}
\newcommand{\NIKHEF}{Nikhef, Science Park 105, 1098 XG Amsterdam, The Netherlands.}

\newcommand{\IWF}{Space Research Institute, Austrian Academy of Sciences, Schmiedlstrasse 6, 8042 Graz, Austria.}
\begin{document}
\title{Predicting Steady-State Behavior in Complex Networks with Graph Neural Networks}





\author{Priodyuti Pradhan} 
\email{prio@iiitr.ac.in} \affiliation{\IIIR}
\author{Amit Reza} 
\email{amit.reza@oeaw.ac.at} \affiliation{\IWF} \affiliation{\NIKHEF}
\begin{abstract}
In complex systems, information propagation can be defined as diffused or delocalized, weakly localized, and strongly localized. This study investigates the application of graph neural network models to learn the behavior of a linear dynamical system on networks. A graph convolution and attention-based neural network framework has been developed to identify the steady-state behavior of the linear dynamical system. We reveal that our trained model distinguishes the different states with high accuracy. Furthermore, we have evaluated model performance with real-world data. In addition, to understand the explainability of our model, we provide an analytical derivation for the forward and backward propagation of our framework.
\end{abstract}

\maketitle
\section{Introduction}
\label{Sec:Intro}
Relations or interactions are ubiquitous, whether the interaction of power grid generators to provide proper functioning of the power supply over a country, or interactions of bio-molecules inside the cell to the proper functioning of cellular activity, or interactions of neurons inside brains to perform specific functions or interactions among satellites to provide accurate GPS services or interactions among quantum particle enabling quantum communications or the recent coronavirus spread ~\cite{revStrogatz2001, dynamicreconfig2011, quantuminternet, GPSnetwork, mahapatra2020multilevel}. All these systems share two fundamental characteristics: a network structure and information propagation among their components.

\begin{figure*}[tbh]
\begin{center}
\includegraphics[width = 6.8in, height = 2.6in]{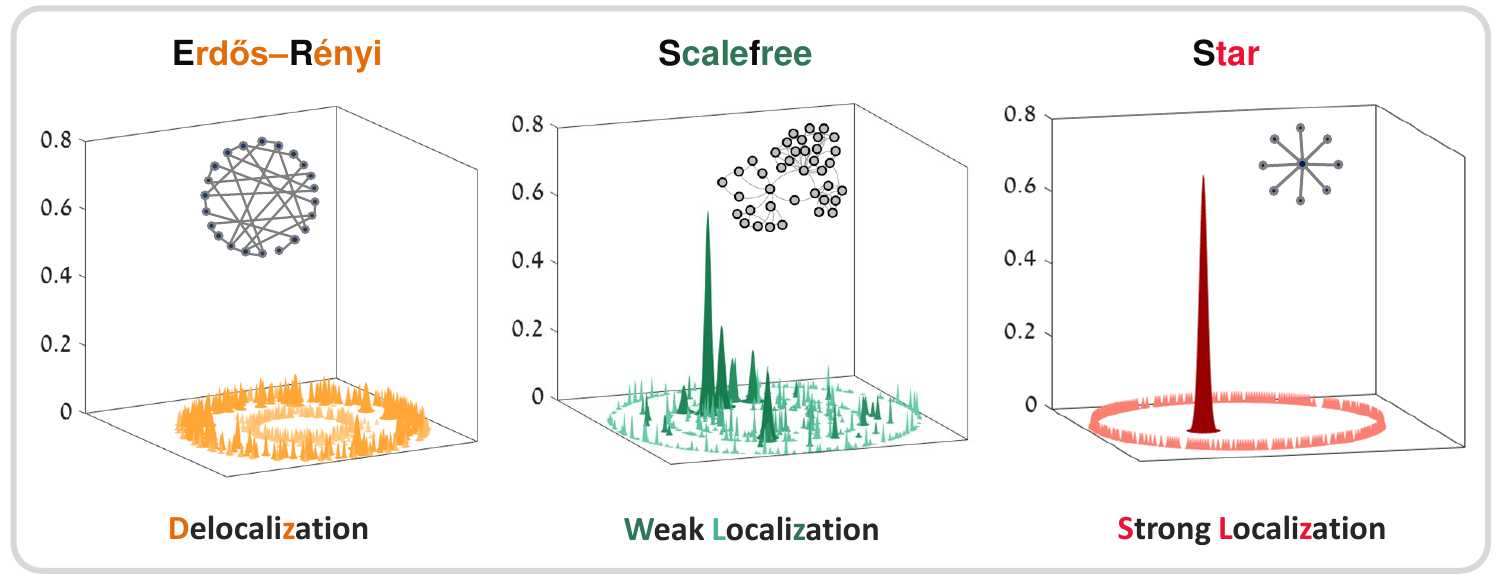}
\caption{Steady-state behavior in linear dynamical systems on complex networks. We depict the nodes in the graphs with $x-y$ coordinates. We assign the sizes of a node based on the degree of a node. The $z-axis$ portrays the amount of information ($x_i^{*}$) on a node in the steady state. The steady-state behavior of linear dynamics on the ER random network leads to delocalization, the scale-free network shows weak localization, and the star network shows a strong localization.}
\label{different_linear_dynamic_states}
\end{center}
\end{figure*}

In complex networks, information propagation can occur in three distinct states - diffused or delocalized, weakly localized, and strongly localized \cite{filoche2012universal}. Localization refers to the tendency of information to condense in a single component (strong localization) or a few components (weak localization) of the network instead of information diffusing evenly (delocalization) throughout the network (Fig. \ref{different_linear_dynamic_states}). Localization or lack of it is particularly significant in solid-state physics and quantum chemistry \cite{elsner1999anderson}, where the presence or absence of localization influences the properties of molecules and materials. For example, electrons are delocalized in metals, while in insulators, they are localized \cite{elsner1999anderson}.

Investigation of (de)localization behavior of complex networks is important for gaining insight into fundamental network problems such as network centrality measure \cite{pradhan2020principal}, spectral partitioning \cite{zhang2016robust}, development of approximation algorithms \cite{gleich2015using}. Additionally, it plays a vital role in understanding a wide range of diffusion processes, like criticality in brain networks, epidemic spread, and rumor propagation \cite{loc2020spectra, revdynamicalprocess2012}. These dynamic processes have an impact on how different complex systems evolve or behave \cite{revdynamicalprocess2012}. For example, understanding epidemic spread can help in developing strategies to slow its initial transmission, allowing time for vaccine development and deployment \cite{diseasespreading2004, pandemicinfluenza2005, urbanizationhantavirus2018, urbanizationinfluenza2018, jalan2020wheel}. The interactions within real-world complex systems are often nonlinear \cite{strogatz2018nonlinear}. In some cases, nonlinear systems can be solved by transforming them into linear systems through changing variables. Furthermore, the behavior of nonlinear systems can frequently be approximated by their linear counterparts near fixed points. Therefore, understanding linear systems and their solutions is an essential first step toward comprehending more complex nonlinear dynamical systems \cite{strogatz2018nonlinear}.

Here, we develop a Graph Neural Network (GNN) architecture to identify the behavior of linear dynamical states on complex networks. We create datasets where the training labels are derived from the inverse participation ratio (IPR) value of the principal eigenvector (PEV) of the network matrices. The GNN model takes the network structure as input and predicts IPR values, enabling the identification of graphs into their respective linear dynamical states. Our model performs well in identifying different states and is particularly effective across varying-sized networks. A key advantage of using GNN is its ability to train on smaller networks and generalize well to larger ones during testing. We also provide an analytical framework to understand the explainability of our model. Finally, we use real-world data sets in our model.

\section{Problem Definition}
We consider a linear dynamical process, $\mathcal{D}$ takes place on unknown network structures represented as $\mathcal{G} = \{V, E\}$ where $V = \{v_{1}, v_{2}, \ldots, v_{n} \}$ is the set of vertices (nodes), $E = \{(v_{i}, v_{j}) | v_{i}, v_{j} \in V \}$ is the set of edges (connections). The degree of a node $i$ in an unweighted graph is the number of nodes adjacent to it, which is given by $\sum_{j=1}^{n} a_{ij}$ where $a_{ij}$ is the adjacency matrix element. The links in $\mathcal{G}$ represent dynamic interactions whose nature depends on context. For instance, in a social system, $a_{ij} = 1$ captures a potentially infectious interaction between individuals $i$ and $j$ \cite{pevecnatphys2013}, whereas, in a rumor-propagation network, it may reflect a human interaction for spreading information. To account for these dynamic distinctions, we denote the activity of each node as $x_i(t)$, which quantifies individual $i$’s probability of infection or rumor propagation. We can track the linear dynamics of node $i$ via 
\begin{equation} 
\begin{split}
\frac{dx_i(t)}{dt}&= \alpha x_i(t)+\beta \sum_{j=1}^{n}a_{ij}x_j(t)
\end{split}
\label{Eq1:power_iteration}
\end{equation}
where $x_i(t)$ is the self-dynamic term, the second term captures the neighboring interactions at time $t$, and $\alpha$, $\beta$ are the model parameters of the linear dynamical system. In matrix notation, we can express Eq. (\ref{Eq1:power_iteration}) as 
\begin{equation}
\frac{d\bm{x}(t)}{dt} =  {\bf M} \bm{x}(t)
\label{Eq2:power_iteration}
\end{equation}
where $\bm{x}(t)=(x_1(t),x_2(t),\ldots,x_n(t))^{T}$, ${\bf M}=\alpha {\bf I}+\beta {\bf A}$ is the transition, ${\bf A}$ is the adjacency, and {\bf I} is the identity matrices, respectively. If $\bm{x}(0)$ is the initial state of the system, the long-term (steady state) behavior ($\bm{x}^{*}$) of the linear dynamical system can be found as 
\begin{equation}
\begin{split}
\bm{x}(t) &= e^{{\bf M}t}\bm{x}(0)\overset{t\rightarrow \infty}{\Longrightarrow} \bm{x}^{*} \sim  \bm{u}_1^{\bf M}
\end{split}
\end{equation}
where $\bm{u}_1^{\bf M}$ is the PEV of {\bf M} (Appendix \ref{linear_dynamics_calculations}). Further, if we multiply both side of ${\bf M} = \alpha {\bf I}+\beta {\bf A}$ by eigenvectors of ${\bf A}$ i.e., $\bm{u}_i^{\bf A}$, we get 
\begin{equation}\nonumber
{\bf M}\bm{u}_i^{\bf A} =[\alpha+\beta\lambda_i^{\bf A}]\bm{u}_i^{\bf A} =\lambda_i^{\bf M}\bm{u}_i^{\bf A}
\end{equation}
We can observe that eigenvectors of {\bf M} are the same as eigenvectors of {\bf A} where $\lambda_i^{\bf M} = \alpha + \beta \lambda_i^{\bf A}$ \cite{loc2020spectra}. Thus,
\begin{equation}\nonumber
\bm{x}^{*} \sim \bm{u}_1^{\bf M} \equiv \bm{u}_1^{\bf A} 
\end{equation}
{\em Therefore, understanding the long-term behavior of the information flow pattern for linear dynamical systems is enough to understand the behavior of PEV of the adjacency matrix.} Further, the behavior of PEV for an adjacency matrix depends on the structure of the network (${\bf A=U \Lambda U^T}$). Hence, we study the relationship between network structure and the behavior of PEV, leading to understanding the behavior of the steady state of linear dynamics.
\begin{figure*}[tbh]
\begin{center}
\includegraphics[width = 6in, height = 2in]{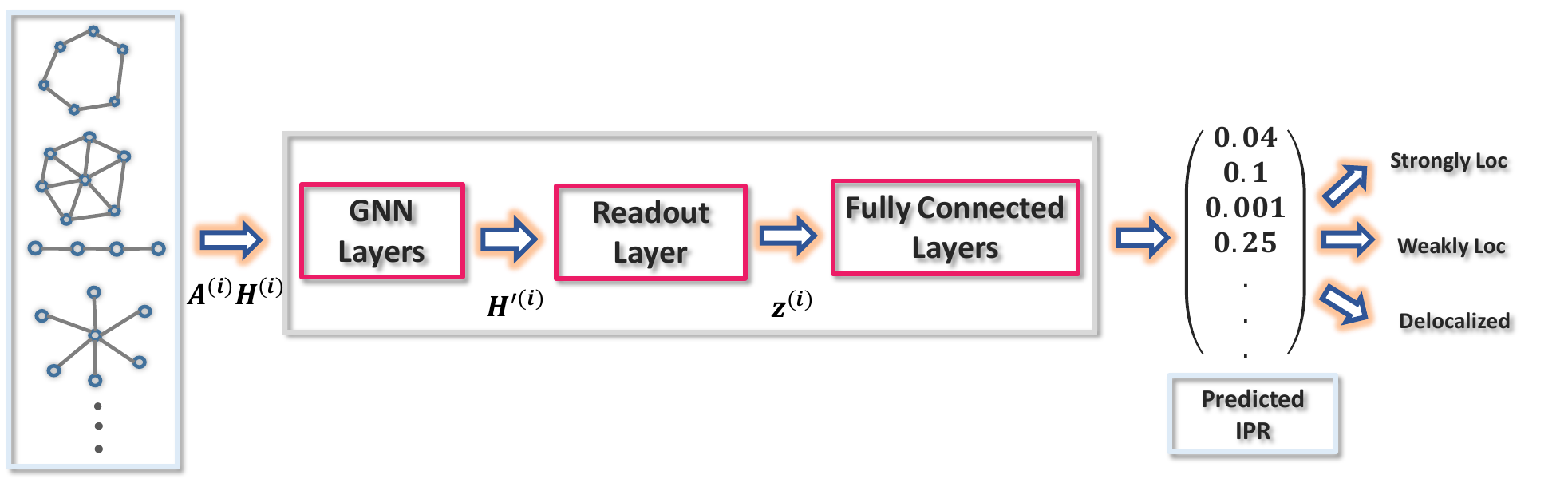}
\caption{The architecture of the Graph Neural Network (GNN) for the regression task over the graphs. The $i^{th}$ input graph (${\bf A }^{(i)}$) and the associated node features (${\bf H}^{(i)}$) are given in matrix form to the models. The GNN layers output an updated node feature matrix ${\bf H}^{'(i)}$, which is passed through a readout layer to produce a graph-level representation $\bm{z}^{(i)}$. A fully connected layer then predicts the IPR value. Finally, we apply a threshold scheme (Eq. (\ref{threshold_scheme})) to identify different linear dynamical states.} 
\label{schematic_NN}
\end{center}
\end{figure*}
We quantify the (de)localization behavior of the steady state ($\bm{x}^{*}$) or the PEV ($\bm{u} \equiv \bm{u}_1^{\mathbf{A}}$) using the inverse participation ratio (IPR), defined as the sum of the fourth powers of the state vector entries \cite{loc2017optimized} as
\begin{equation} \label{eq_IPR}
y_{\bm{x}^{*}} = \frac{\sum_{i = 1}^{n} {x^{*}_{i}}^{4}}{\biggl[\sum_{i = 1}^{n} {x^{*}_{i}}^{2}\biggr]^{2}} 
\end{equation}
where $x^{*}_{i}$ is the $i^{\text{th}}$ component of $\bm{x^{*}}=(x^{*}_1,x^{*}_2,\dots,x^{*}_n)^{T}$ and $\sum_{i = 1}^{n} {x^{*}_{i}}^{2}$ is the normalization term. A vector with component $(c,c,\ldots,c)^{T}$ is delocalized and has $y_{\bm{x}^{*}} = \frac{1}{n}$ for some positive constant $c>0$, whereas the vector with components $(1, 0, \ldots, 0)^{T}$ yields $y_{\bm{x}^{*}} = 1$ and referred as most localized. Furthermore, we consider the networks to be simple, connected and undirected {\bf \cite{mieghambook2011}}. Hence, some information can easily propagate from one node to another and we never get a steady-state vector of the form $\bm{x}^{*} = (1, 0, \ldots, 0)^{T}$ for a connected network, and thus the IPR value lies between $ \frac{1}{n} \leq y_{\bm{x}^{*}} < 1$. Therefore, the localization-delocalization behavior of the linear dynamics in the network is quantified using a real value, i.e., each graph associates an IPR value \cite{pradhan2018network,loc2017optimized, pradhan2020principal, loc2020spectra}. We also employ entropy measurement for the localization, and more details are in Appendix \ref{entropy_in_pev}. To identify the states ($\bm{x}^{*}$) belong to which category of dynamical behavior for linear dynamics, we formalize a threshold scheme for identifying IPR values ($y \equiv y_{\bm{x}^{*}}$) lies in the range $[1/n, 1)$. We define two thresholds, $\tau_1$ and $\tau_2$, such that $1/n \leq \tau_1 < \tau_2 < 1$. An additional parameter $\epsilon$ ($\epsilon > 0$) defines some flexibility around the thresholds.

\vspace{2mm}
\noindent {{\em \textbf{Delocalized region}} ($r_{1}$).} IPR values significantly below the first threshold, including an $\epsilon$-width around $\tau_1$:
\begin{equation}\nonumber
r_1 = \{y \in [1/n, 1) \mid y \leq \tau_1 - \epsilon \}    
\end{equation}
\noindent {{\em \textbf{Weakly localized region}} ($r_{2}$).} IPR values around and between the two thresholds, including $\epsilon$-width around $\tau_{1}$ and $\tau_{2}$:
\begin{equation}\nonumber
r_{2} = \{y \in [1/n, 1) \mid \tau_1 - \epsilon < y < \tau_{2} + \epsilon \}
\end{equation}
\noindent {{\em \textbf{Strongly localized region}} ($r_{3}$).} 
IPR values significantly above the second threshold, including an $\epsilon$-width around $\tau_{2}$:
\begin{equation}\nonumber
r_{3} = \{y \in [1/n, 1) \mid y \geq \tau_{2} + \epsilon \}
\end{equation}   
The regions can be defined using a piece-wise function:
\begin{equation} 
\label{threshold_scheme}
r(y, \tau_1, \tau_2, \epsilon) =
\begin{cases}
1 & \text{if } y \leq \tau_1 - \epsilon \\
2 & \text{if } \tau_1 - \epsilon < y < \tau_2 + \epsilon \\
3 & \text{if } y \geq \tau_2 + \epsilon \\
\end{cases}
\end{equation}   
For instance, we consider a set of threshold values as $\tau_1 = 0.05$, $\tau_2 = 0.2$, $\epsilon = 1e-6$. Now, if we consider a regular network (each node have the same degree) of $n$ nodes, we have PEV, $\bm{u}^{\mathcal{R}}=(\frac{1}{\sqrt{n}},\frac{1}{\sqrt{n}},\ldots,\frac{1}{\sqrt{n}})^{T}$ of {\bf A} (Theorem 6 \cite{mieghambook2011}) yielding, $y_{\bm{u}^{\mathcal{R}}}=\frac{1}{n}$, thus $y_{\bm{u}^{\mathcal{R}}} \rightarrow0$ as $n\rightarrow \infty$. On the other hand, for a star graph having $n$ nodes, $\bm{u}^{\mathcal{S}}=\biggl(\frac{1}{\sqrt{2}},\frac{1}{\sqrt{2(n-1)}},\ldots, \frac{1}{\sqrt{2(n-1)}}\biggr)^{T}$ and, $y_{\bm{u}^{\mathcal{S}}} = \frac{1}{4} + \frac{1}{4(n-1)}$. Hence, for $n \rightarrow \infty$, we get $y_{\bm{u}^{\mathcal{S}}} 
\approx 0.25$, and PEV is strongly localized for the star networks. Further, it is also difficult to find a closed functional form of PEV for any network, and thereby, it is hard to find the IPR value analytically. For instance, in Erd\"os-R\'enyi (ER) random networks, we get a delocalized PEV due to each node having the same expected degree \cite{delocpev}. In contrast, the presence of power-law degree distribution for SF networks leads to some localization in the PEV. For SF networks, the IPR value, while being larger than the ER random networks, is much lesser than the star networks \cite{Goltsevprl2012}. It may seem that when the network structure is close to regular, linear dynamics are delocalized, and increasing degree heterogeneity increases the localization. However, always looking at the degree heterogeneity not able to decide localization and analyzing structural and spectral properties is essential \cite{loc2017optimized, loc2020spectra}. A fundamental question at the core of the structural-dynamical relation is: Can we predict the steady state behavior of a linear dynamical process in complex networks? 

\begin{figure*}[tbh]
\begin{center}
\includegraphics[width = 6.5in, height = 3.6in]{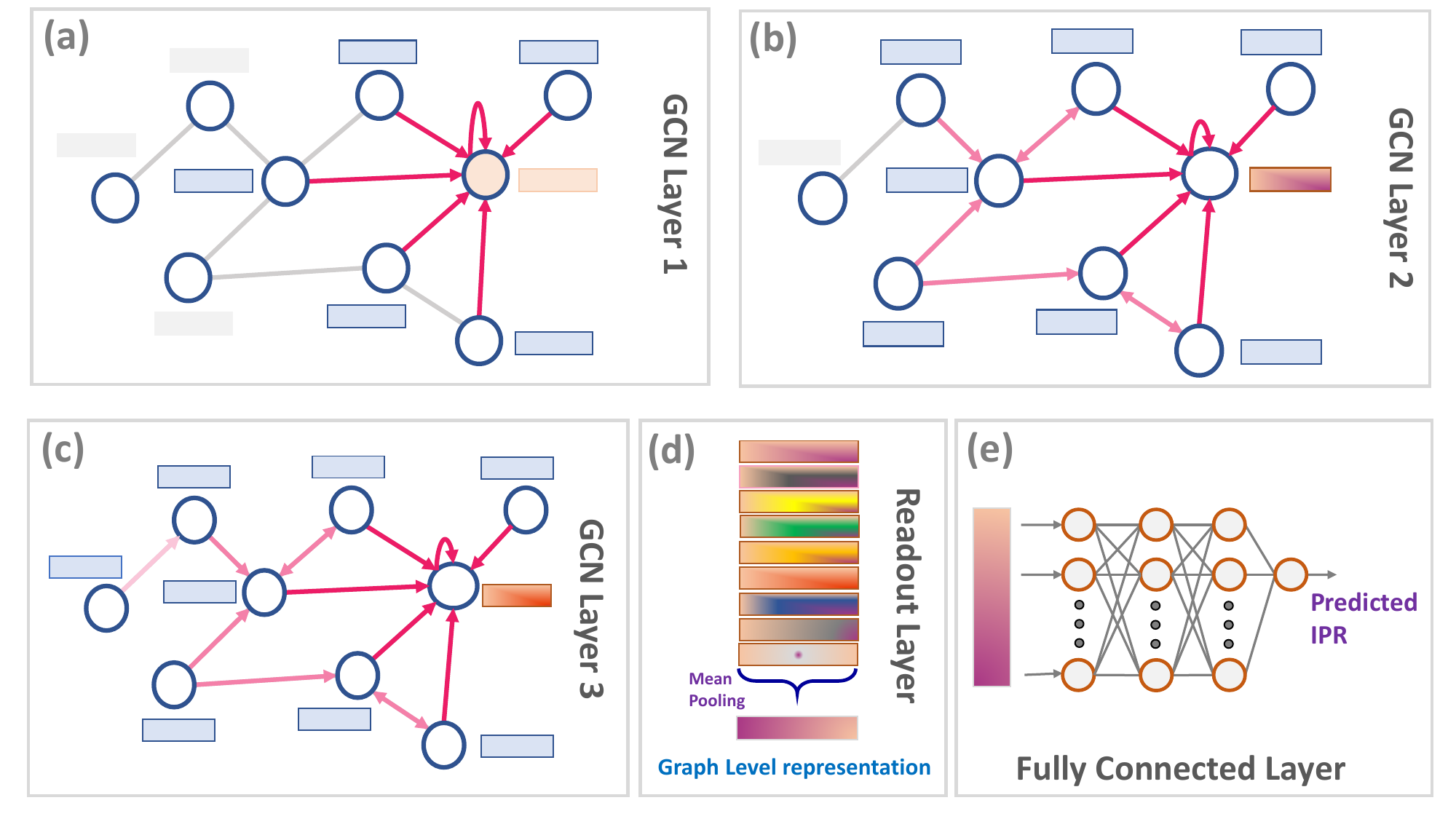}
\caption{Neural message passing in Graph Convolutional Networks (GCN) model for the regression task. Each node's features are represented with a small rectangle associated with the node. (a) represents one layer of GCN where a node aggregates and updates its feature based on the immediate neighbor and its features. (b) For the second GCN layer, a node updates its features by aggregating the messages from neighbors to neighbors and its own. (c) The third layer of GCN aggregates the messages from neighbors to neighbors to neighbors. (d) The readout layer creates a graph-level representation from all node features through the mean pooling function that holds the graph's global information based on the neural message passing framework. (e) Finally, we pass it to a fully connected neural network for the graph's IPR value prediction task.} 
\label{message_passing_GNN}
\end{center}
\end{figure*}
\begin{figure*}[tbh]
\begin{center}
\includegraphics[width = 5.6in, height = 3.8in]{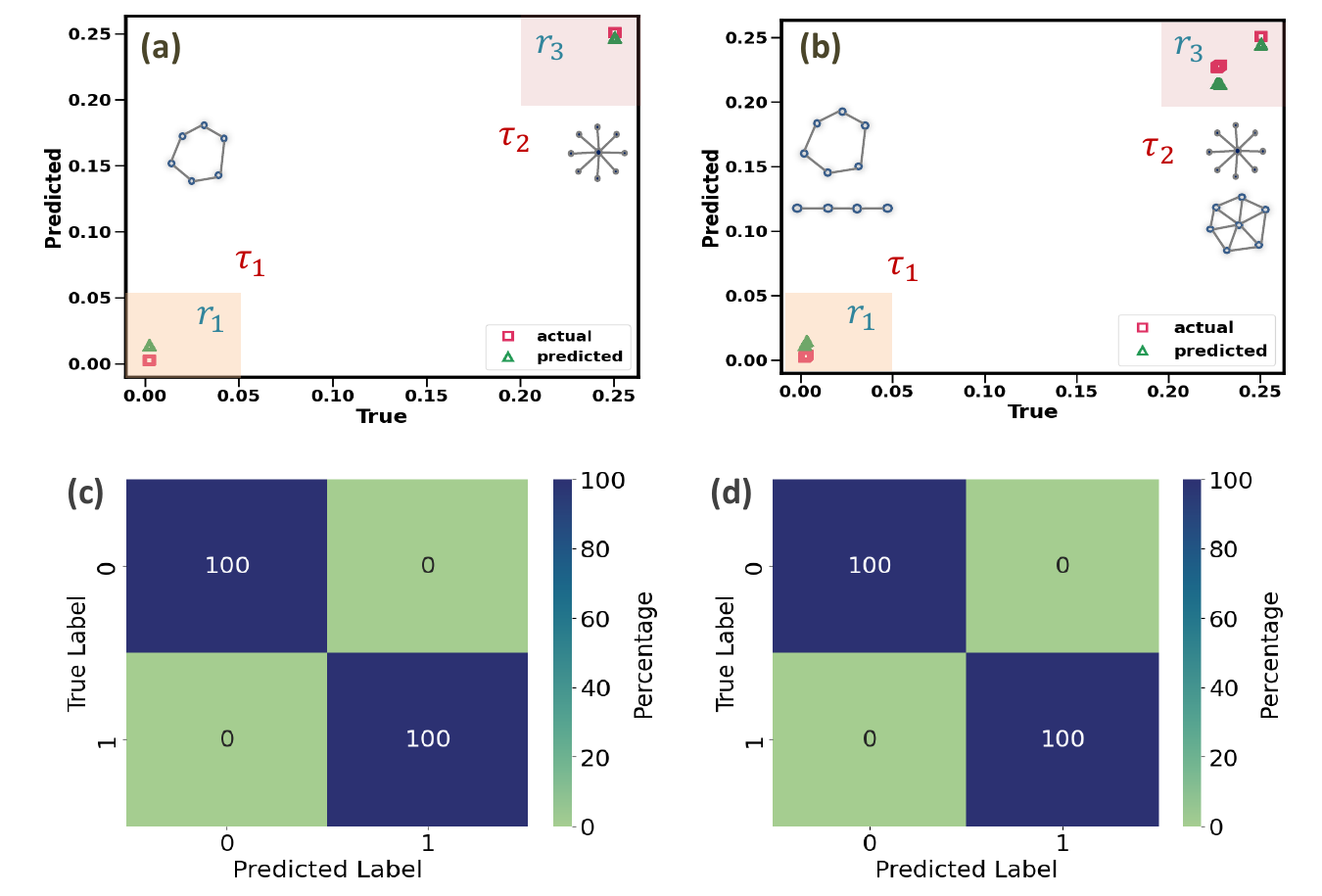}
\caption{We train the GCN model with two types of structures associated with delocalized and strongly localized states. The input to the GCN model is the cycle and star graphs and the associated target values, i.e., IPR. (a) We give datasets with varying-sized cycle and star networks during the test time. We can observe the true and predicted IPR values. Finally, we apply the threshold function (Eq. (\ref{threshold_scheme})) on the predicted IPR value. For our study, we choose the parameters for the threshold values as $\tau_{1} = 0.05$, $\tau_{2} = 0.2$, and $\epsilon = 1e-6$ in Eq. (\ref{threshold_scheme}). We mark the delocalized region ($r_1$) with an orange color box and the strongly localized region ($r_{3}$) with a red color box based on $\tau_1$ and $\tau_2$ values. This visualization enables us to identify the test network as delocalized and strongly localized based on their predicted IPR values. If predicted IPR values fall within the designated region, the identification of steady-state behavior prediction is correct. Notably, the threshold scheme also allows us to identify the correct behavior even when predicted IPR values deviate from the original values but lie within the threshold boundary. (b) To observe the expressivity of the model, we incorporate two other different network structures (wheel and path networks) and repeat the process. (c, d) We can observe the model-predicted values with high accuracy in the confusion matrix (in $\%$). } 
\label{loc_deloc_results_undirected}
\end{center}
\end{figure*}
Here, we formulate the problem as a graph regression task to predict a target value, IPR, associated with each graph structure. For a given set of graphs $\{\mathcal{G}_i\}_{i=1}^N$ where each  $\mathcal{G}_i = (V_{i}, E_{i})$ consists of a set of nodes $V_i$ and a set of edges $E_{i}$ such that $n_i=|V_i|$ and $m_i=|E_i|$. We represent each $\mathcal{G}_i$ using its adjacency matrix ${\bf A}^{(i)}$. Further, each graph $\mathcal{G}_i$ has an associated target value, i.e., IPR value, $y^{(i)} \in \mathbb{R}$. For each node $v \in V_{i}$ in $ \mathcal{G}_i$, there is an associated feature vector $\bm{h}_j^{(i)} \in \mathbb{R}^{1 \times d}$ and $\mathbf{H}^{(i)} \in \mathbb{R}^{|V_i| \times d}$ be the node feature matrix where $\bm{h}_j^{(i)}$ is the $j^{th}$ row. The objective is to learn a function $f: \mathcal{G} \to \mathbb{R}$, such that $f(\mathcal{G}_i) \approx y^{(i)}$ for the given set of $N$ graphs.

\section{Methodology and Results}
The function $f$ can be parameterized by a model, in our case, Graph Convolutional Networks (GCN) and Graph Attention Networks (GAT) (Appendices \ref{GCN_math_details}, \ref{GAT_math_details}). Let $\bm{\theta}$ be the parameters of the model. The prediction for $\mathcal{G}_i$ is denoted by $ \hat{y}^{(i)} = f(\mathcal{G}_i; \bm{\theta})$. The model parameters $\bm{\theta}$ are learned by minimizing a loss function that measures the difference between the predicted values $\hat{y}^{(i)}$ and the true target values $y^{(i)}$. We use the Mean Squared Error (MSE) loss function as
\begin{equation}
\mathcal{L}(\bm{\theta}) = \frac{1}{N} \sum_{i=1}^N (\hat{y}^{(i)} - y^{(i)})^2 
\end{equation}
To improve numerical stability, we apply a logarithmic transformation and compute the loss as
\begin{equation}\label{log_MSE}
\mathcal{L} = \text{MSE}(\log(1 + \hat{y}), \log(1 + y))
\end{equation}
This formulation helps compress the scale of the outputs and targets, particularly useful when the target values span multiple orders of magnitude. Hence, the graph regression problem can be formalized as finding the optimal parameters $\bm{\theta}$ of a model $f$ that minimize the loss function $\mathcal{L}(\bm{\theta})$, enabling the prediction of IPR values for given graphs. 
\subsection{GCN Architecture}

The GCN architecture comprises three graph convolutional layers, each followed by a ReLU activation function. After that, a readout layer performs mean pooling to aggregate node features into a single graph representation. Finally, we use a fully connected layer that outputs the scalar IPR value for the regression task (Fig. \ref{schematic_NN}). A brief description of the architecture is provided below.

\vspace{2mm}
\noindent {\bf Input Layer:}
The input layer recieves a normalized adjacency ($\hat{\mathbf{A}}$) and initial node feature \({\bf H}^{(0)}\) matrices.

\vspace{2mm}
\noindent {\bf Graph Convolution Layers:}
We stack three graph convolution layers (Eq. \ref{gcn_layers}) to capture local and higher-order neighborhood information of a node. After each graph convolution layer, we apply nonlinear activation functions (ReLU). Each layer uses the node representations from the previous layer to compute updated representations in the current layer. The first layer of GCN facilitates information flow between first-order neighbors (Fig. \ref{message_passing_GNN}(a)), while the second layer aggregates information from the second-order neighbors, i.e., the neighbors of a node's neighbors (Fig. \ref{message_passing_GNN}(b)), and this process continues for subsequent layers (Fig. \ref{message_passing_GNN}(c)) and we get 
\begin{equation}
\label{gcn_layers}
\mathbf{H}^{(l)} = \sigma \big( \hat{\mathbf{A}} {\bf H}^{(l-1)} \mathbf{W}^{(l-1)} \big),\; l = \{1, 2, 3\}
\end{equation}
where ${\bf H}^{(0)} \in \mathbb{R}^{n \times d}$ is the initial input feature matrix, and $\mathbf{W}^{(0)} \in \mathbb{R}^{d \times k_0}, \mathbf{W}^{(1)} \in \mathbb{R}^{k_0 \times k_1}, \mathbf{W}^{(2)} \in \mathbb{R}^{k_1 \times k_2}$ are the weight matrices for the first, second, and third layers, respectively. Hence, ${\bf H}^{(1)} \in \mathbb{R}^{n \times k_0}$, and ${\bf H}^{(2)} \in \mathbb{R}^{n \times k_1}$ are the intermediate node features matrices and after three layers of graph convolution, final output node features are represented as $\mathbf{H}^{(3)} \in \mathbb{R}^{n \times k_2}$.

\begin{figure*}[tbh]
\begin{center}
\includegraphics[width = 6.5in, height = 1.6in]{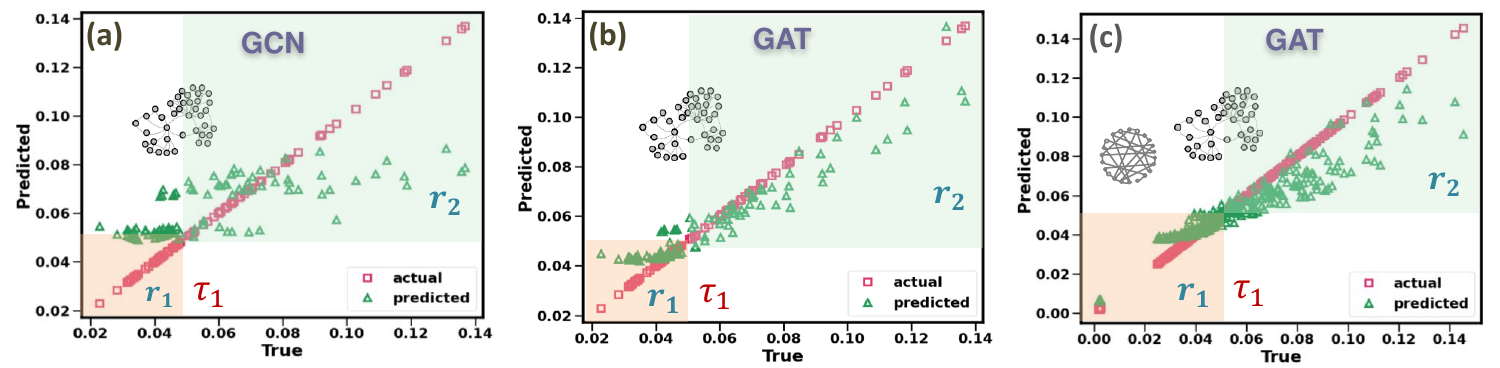}
\caption{We use scale-free networks for training and testing. (a) We can observe very low accuracy during the test time for the GCN model. (b) However, we can observe increased accuracy using the Graph Attention Network. (c) We consider ER random and scale-free networks and associated IPR values leading to the dynamical states being delocalized and weakly localized. We can observe the GAT model predicts the state's IPR value with significant accuracy. Here, we train the model for $200$ epochs.}
\label{loc_deloc_results_undirected_multi}
\end{center}
\end{figure*}

\vspace{2mm}
\noindent {\bf Readout Layer:}
For the scalar value regression task over a set of graphs, we incorporate a readout layer to aggregate all node features into a single graph-level representation (Fig. \ref{message_passing_GNN}(d)). We use mean pooling as a readout function, 
\begin{equation}
\nonumber
\bm{z} = \texttt{READOUT}(\mathbf{H}^{(3)})= \frac{1}{n} \sum_{j=1}^{n} \bm{h}^{(3)}_j 
\end{equation}
where $\bm{h}^{(3)}_j \in \mathbb{R}^{1 \times k_2}$ is the $j^{th}$ row of ${\bf H}^{(3)} \in \mathbb{R}^{n \times k_2}$ and $\bm{z} \in \mathbb{R}^{1 \times k_2}$. Finally, we pass it to a linear part of the basic neural network (Fig. \ref{message_passing_GNN}(e)) to perform regression task over graphs and output the predicted IPR value as  
\begin{equation}\nonumber
\hat{y} = \bm{z} \mathbf{W}^{\text{(lin)}}+b    
\end{equation}
where $\mathbf{W}^{\text{(lin)}} \in \mathbb{R}^{k_2 \times 1}$ is the weight matrix and $b \in \mathbb{R}$ is the bias value. For our architecture, $\bm{\theta}=\{ \mathbf{W}^{(0)},\mathbf{W}^{(1)},\mathbf{W}^{(2)},\mathbf{W}^{\text{(lin)}}, b \}$. After getting the predicted IPR value, we use the threshold scheme (Eq. \ref{threshold_scheme}) to identify the steady state behavior on complex networks.
\subsection{Data Sets Preparation}
For the regression task, we create the graph data sets by combining delocalized, weakly localized, and strongly localized network structures, which include star, wheel, path, cycle, and random graph models as Erd\H{o}s-R\'enyi (ER) and the scale-free (SF) networks (Appendix \ref{complex_networks}). As predictions of the GNN model are independent of network size, we vary the size of the networks during dataset creation and store them as adjacency matrices (${\bf A}^{(i)}$). For each network, we calculate the IPR value from the principal eigenvector (PEV) and assign it as the target value ($y^{(i)}$) to $\mathcal{G}_i$ in the datasets. Since we do not have predefined node features for the networks, and the GCN framework requires node features as input \cite{hamilton2020graph}, we initialize the feature vector ($\bm{h}_j^{(i)}$) for each node with network-specific properties (clustering coefficient, PageRank, degree centrality, betweenness centrality and closeness centrality) to form the initial feature matrix (${\bf H}^{(0, i)}$) for the $i$th graph. However, the feature matrix could also be initialized with random binary values (i.e., $0$, $1$) \cite{hamilton2020graph} (Appendix \ref{random_vs_network_features}). We pass the adjacency matrices, feature matrices, and labels into the model for the regression task, $\{({\bf A}^{(i)},{\bf H}^{(0,i)}, y^{(i)})\}_{i=1}^{N}$. We primarily use small-sized networks for training the model, while testing is conducted on networks ranging from small to large sizes, including sizes similar to and beyond those used in training. 

We also consider real-world benchmark datasets ($\texttt{ENZYMES}$, $\texttt{NCI}1$ and $\texttt{MCF}-7$) to train the model \cite{ivanov2019understanding, morris2020tudataset}. $\texttt{ENZYMES}$ is a dataset of $\texttt{N}_\texttt{ENZ} = 600$ protein tertiary structures obtained from the $\texttt{BRENDA}$ enzyme database. The $\texttt{NCI}1$ graph dataset is a benchmark dataset used in cheminformatics, where each graph represents a chemical compound, with nodes representing atoms and edges representing bonds between them. The $\texttt{NCI}1$ contains $\texttt{N}_{\texttt{NCI}} = 4110$ graphs, and each node has $37$ features. The $\texttt{MCF}-7$ dataset consists of small molecule activities against breast cancer tumors of $\texttt{N}_{\texttt{MCF}} = 27770$ graphs where each node has $46$ features. We access the datasets through $\texttt{PyTorch}$ Geometric libraries and preprocess the data sets by removing all the disconnected graphs and those with fewer than $10$ nodes. We incorporate node features extracted from network-specific properties. The sizes of our preprocessed datasets are $\texttt{N}_{\texttt{ENZ}} = 441$, $\texttt{N}_{\texttt{NCI}} = 2796$, and $\texttt{N}_{\texttt{MCF}} = 25084$ with each nodes having network-specific features as in the model network. Note that IPR values for disconnected graphs are trivially high, and we focus exclusively on connected graphs in our study. For real-world data sets, we divide $80\%$ of the data for training and the rest for testing.

\subsection{Training and Testing Strategy}
During the training, the GCN model initializes the model parameters ($\bm{\theta}$). We initialize ${\bf W}^{(l-1)}$ at random using the initialization described in Glorot \& Bengio (2010) \cite{glorot2010understanding}. During the forward pass, for each graph $\mathcal{G}_i$, we compute the graph representation using the GCN layers, which involves message passing and aggregation of node features (Fig. (\ref{message_passing_GNN})). Finally, the model predicts the target value $\hat{y}^{(i)}$. Further, the model computes the loss $\mathcal{L}(\bm{\theta})$ using the predicted ($\hat{y}^{(i)}$) and true target ($y^{(i)}$) values. During the backward pass, the model computes the gradients of the loss with respect to the model parameters. In the next step, the model updates the parameters using an optimization algorithm such as Adam with a learning rate of $0.01$ and a weight decay of $5e-4$. We repeat the forward propagation, loss computation, backward propagation, and parameter update steps until convergence. For our experiment, we choose weight matrix sizes as $k_0=k_1=k_2=64$ in different layers.

We start by creating a simple experimental setup where the input dataset contains only two different types of model networks. One type of network (cycle) is associated with delocalized steady-state behavior, and another (star) is in the strongly localized behavior. During the training phase, we send the adjacency matrix (${\bf A}^{(i)}$), node feature matrix (${\bf H}^{(0,i)}$), and IPR values ($y^{(i)}$) associated with the graphs as labels for the regression task. Once the model is trained, one can observe that the GCN model accurately predicts the IPR value for the two different types of networks (Fig. \ref{loc_deloc_results_undirected}(a)). More importantly, we train the model with smaller-size networks ($n_i=200$ to $300$) and test it with large-size networks ($n_i=400$ to $500$) and training datasets contains $N_{train}=1000$ networks and testing datasets size as $N_{test}=500$ where $50\%$ are cycle and $50\%$ are star networks. Thus, the training cost would be less, and it can easily handle large networks. Furthermore, for the expressivity of the model, we increase the datasets by incorporating two more different types of graphs (path and wheel graphs), where one is delocalized and the other is in strongly localized structures, and we trained the model. We repeat the process by sending the datasets for the regression task to our model and observing that the model provides good accuracy for the test data sets (Fig. \ref{loc_deloc_results_undirected}(b)). Finally, we apply the threshold function (Eq. \ref{threshold_scheme}) on the predicted values and achieve very high accuracy in identifying the dynamic state during the testing (Fig. \ref{loc_deloc_results_undirected}(c, d)). One can observe that the GCN model learns the IPR value well for the above network structures.

We move further and incorporate random graph structures (ER and scale-free random networks) in the data sets. Note that the ER random graph belongs to the delocalized state, and SF belongs to both the delocalized and weakly localized state. We train the model with only the SF networks, and during the testing time, one can observe accuracy is not good (Fig. \ref{loc_deloc_results_undirected_multi}(a)). To resolve this, we changed the model and the parameters. We choose the Graph Attention network \cite{velivckovic2017graph}, update the loss function by considering the $\log$ value (Eq. \ref{log_MSE}), and choose AdamW optimizer instead of Adam. We also use a dropout rate of $0.6$ and set learning rare to $1e-5$ in the model. The new setup leads to improvement in the results (Fig. \ref{loc_deloc_results_undirected_multi}(b)). Now, we consider both ER and SF networks and train the model, and during the testing time, we can observe good accuracy in predicting the IPR values (Fig. \ref{loc_deloc_results_undirected_multi}(c)). We also consider other graph models (Fig. \ref{GCN_IPR_ENTROPY}(a, c)) and entropy measures for localization (Fig. \ref{GCN_IPR_ENTROPY}(b, d)) and observe that our models perform well (Appendices \ref{complex_networks}, \ref{entropy_in_pev}). 
\begin{figure}[tbh!]
\begin{center}
\includegraphics[width = 3.4in, height = 1.2in]{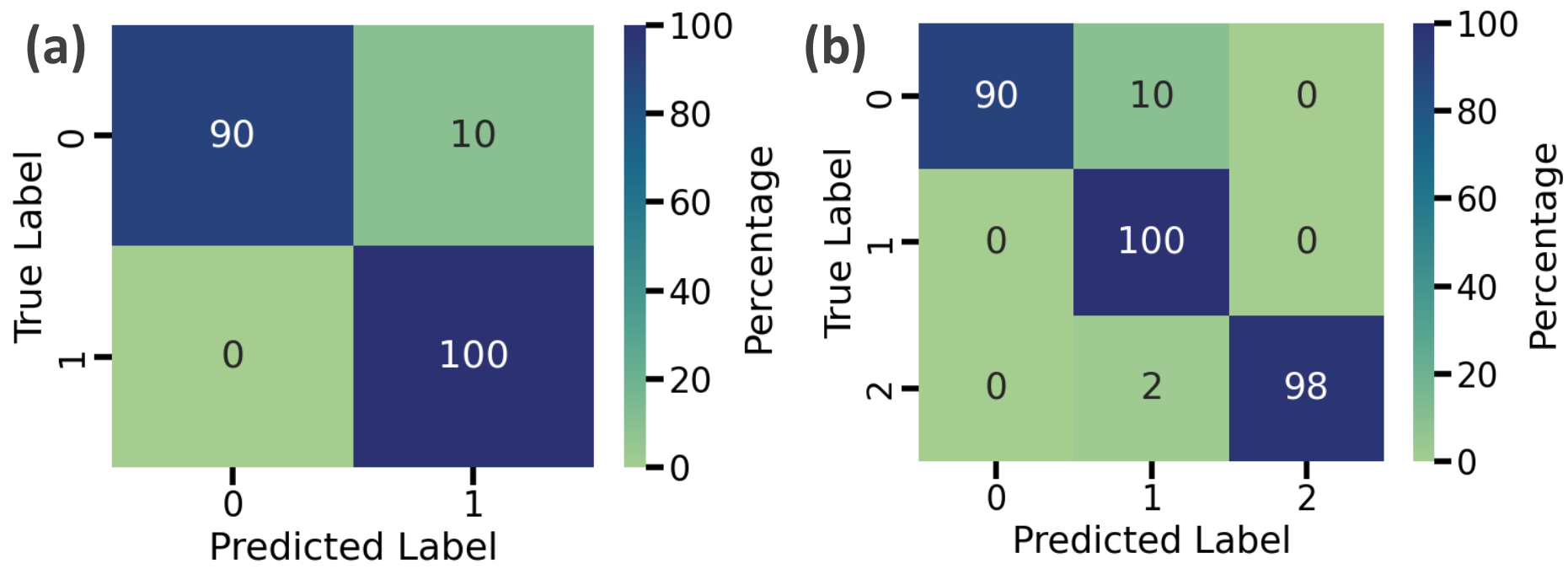}
\caption{Scale invariant. Confusion matrices for the GAT model trained on medium-sized networks (size $500-1500$) and tested on significantly larger networks (size $5000-15000$). (a) Binary classification between delocalized and localized states achieves $95\%$ accuracy. (b) Three-class classification (delocalized ($r_1 \mapsto 0$), weakly localized ($r_2 \mapsto 1$), and strongly localized ($r_3 \mapsto 2$)) achieves $96\%$ accuracy, demonstrating scale-invariant performance of the model.}
\label{fig:cm_large_network_GAT}
\end{center}
\end{figure}

\begin{figure*}[tbh]
\begin{center}
\includegraphics[width = 7.2in, height = 1.7in]{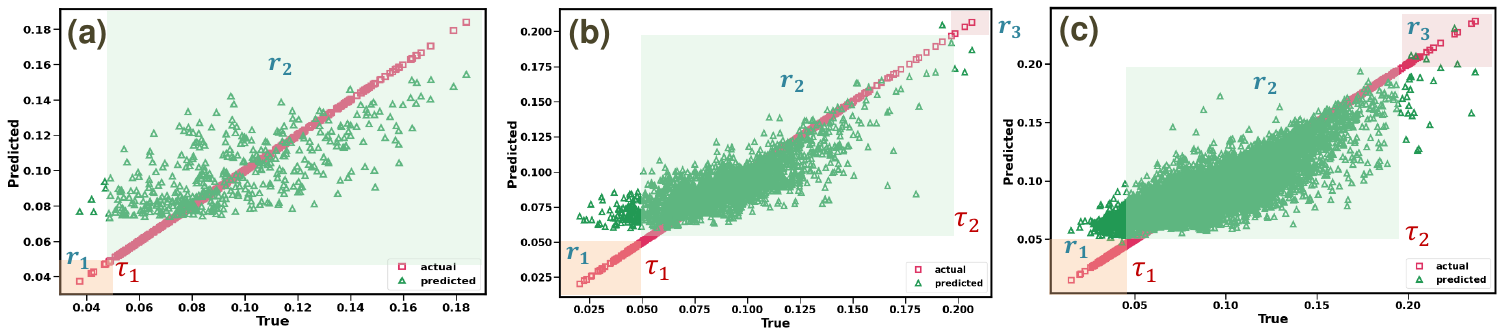}
\caption{Prediction of Graph Attention Networks (GAT) on real world data sets ($\texttt{ENZYMES}$, $\texttt{NCI}1$ and $\texttt{MCF}-7$) \cite{ivanov2019understanding, morris2020tudataset}. (a-c) Our model can also predict the IPR value during testing for real-world graph data sets. We can observe the GAT model predicts the state's IPR value with significant accuracy. (b)  We train the model for $200$ epochs for the real-world dataset. We choose $\tau_{1} = 0.05$, $\tau_{2} = 0.2$, $\epsilon = 1e-6$.}
\label{loc_deloc_results_undirected_real}
\end{center}
\end{figure*}

Further, to understand whether the framework is independent of network size, i.e., ``scale-invariant", we substantiate a more consistent study to evaluate the model's performance on larger graphs. We have trained the GAT model with a network size of $500-1500$, $1000$ for each network type. The trained models have been tested on a network size of $5000 - 15000$, $200$ for each network type, and use the threshold function (Eq. \ref{threshold_scheme}). Fig. \ref{fig:cm_large_network_GAT} represents the confusion matrices corresponding to the binary classification (i.e., delocalized and localized) and three-class classification (i.e., delocalized, weakly localized, and strongly localized)  with $95 \%$ and $96 \%$ accuracy, respectively. Although trained models reliably predict states in model networks, applying them to real-world data presents challenges due to imbalanced state distribution and limited dataset size in the $r_1$ and $r_3$ regions (Fig.~\ref{loc_deloc_results_undirected_real}). 
To assess real-world applicability, we trained the GAT model on real-world data sets and achieved reasonable accuracy on test datasets (Fig.~\ref{loc_deloc_results_undirected_real}(a-c)).

The performance of GCN and GAT in identifying various dynamic states in model networks is highly accurate. GCN is particularly effective in distinguishing between strongly localized and delocalized states (Fig. \ref{loc_deloc_results_undirected}), while GAT excels at differentiating weakly localized and delocalized states (Fig. \ref{loc_deloc_results_undirected_multi}(c)). We observed that the GCN model tends to underperform for specific graph topologies, especially for scale-free and ER random graphs, when tasked with predicting the IPR values. One possible reason is that GCN relies on spectral convolution using normalized adjacency information, which assumes smoothness in node features across the graph \cite{kipf2016semi,zhu2020beyond}. ER networks generally lack sufficient structural regularity \cite{zhu2020beyond}, and scale-free networks exhibit highly skewed degree distributions and strong local heterogeneities. These features can contravene the homophily assumption and reduce the effectiveness of GCN’s neighborhood aggregation. On the contrary, GAT model leverages learnable attention mechanisms that adaptively weight neighbours based on feature relevance rather than uniform or degree-based contributions. This accedes to GAT for better capture of local variations and importance, which could be particularly beneficial for the topologies. While GAT consistently performs better in specific regimes, we prefer to retain both models in our study to provide a comparative understanding of model behavior across diverse graph topologies. It is crucial to highlight that GCN is always faster and easier to understand \cite{wu2020comprehensive} than other models. This dual-model perspective allows GCN to serve as a valuable diagnostic tool to better isolate the influence of topology on model performance. Nevertheless, preferring GAT as the primary model for deployment or end-use scenarios could be a more reasonable and practical solution when robustness across topologies is essential. 

\section{Ablation Study}

\begin{figure}[tbh!]
\begin{center}
\includegraphics[width = 3.4in, height = 2.3in]{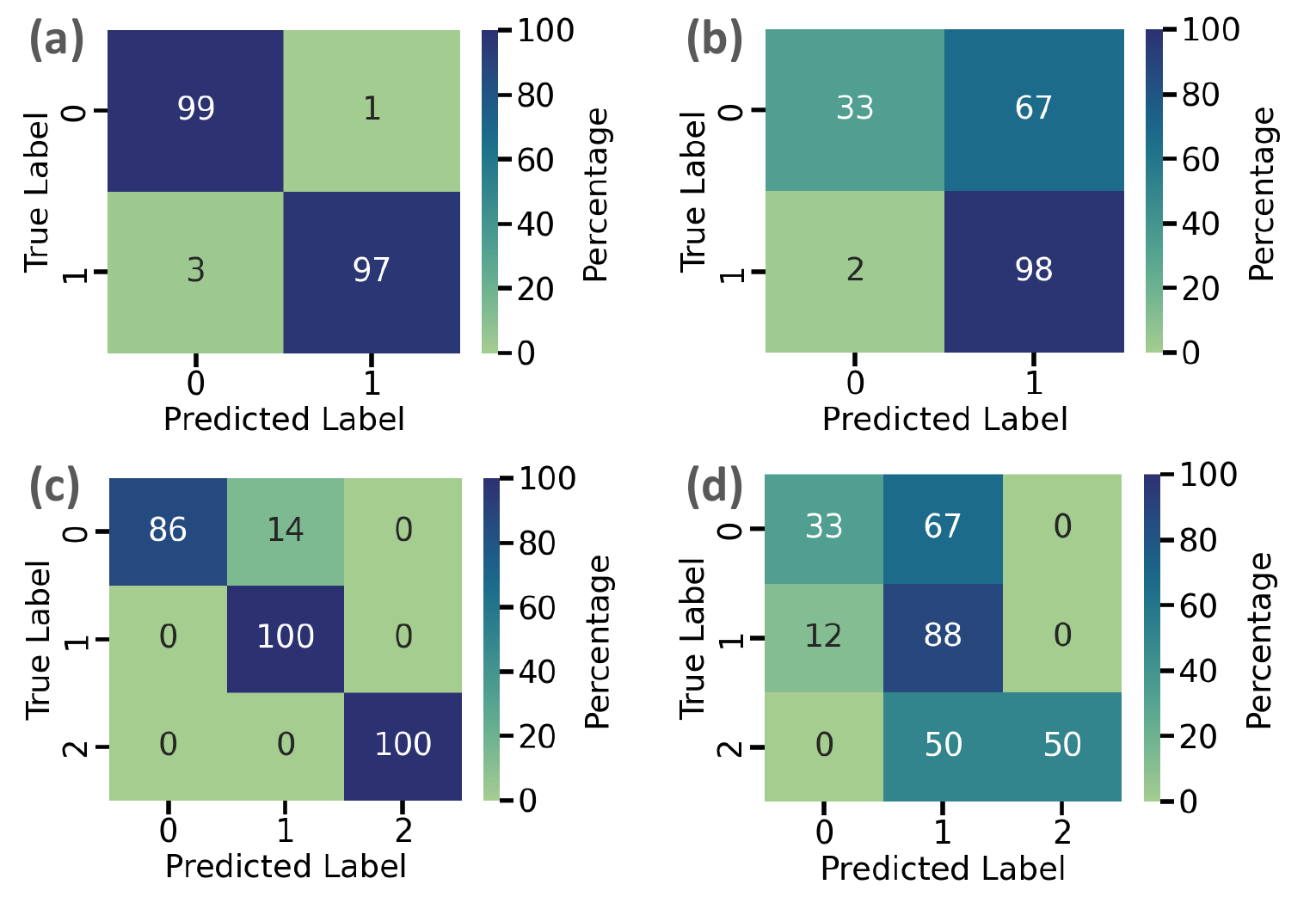}
\caption{Confusion matrices for the GAT model trained with single node-features.
(a, b) Binary classification (strongly localized vs. delocalized) results for betweenness and clustering coefficient features, respectively.
(c, d) Three-class classification (strongly localized, weakly localized, and delocalized) results for closeness and clustering coefficient features, respectively. Betweenness and closeness centralities yield the highest accuracies ($\sim98\%$ and $\sim95.3\%$), while clustering performs the worst in both tasks.}
\label{fig:cm_single_feature}
\end{center}
\end{figure}

To determine the most relevant node features among degree centrality, clustering coefficient, betweenness, closeness, and PageRank, we trained GAT models with individual features separately and compared their accuracy with test data. 
We observed that among the five features, the GAT model trained only on betweenness centrality achieved the best prediction accuracy on the test data (Fig. \ref{fig:cm_single_feature}(a)), while the model using the clustering coefficient yielded the weakest performance in classifying between localized and delocalized states (Fig. \ref{fig:cm_single_feature}(b)). Accuracy is the proportion of correctly predicted classes among the total number of tests made. 
Based on prediction accuracy on the localized and delocalized classification tasks, we can rank the importance of these features as  betweenness ($ \sim 98\%$), closeness ($\sim 93\%$), PageRank ($\sim 90\%$), degree centrality ($\sim 86\%$), and clustering coefficient ($\sim 66\%$).
\begin{figure*}[tbh!]
\begin{center}
\includegraphics[width = 6.8in, height = 4.8in]{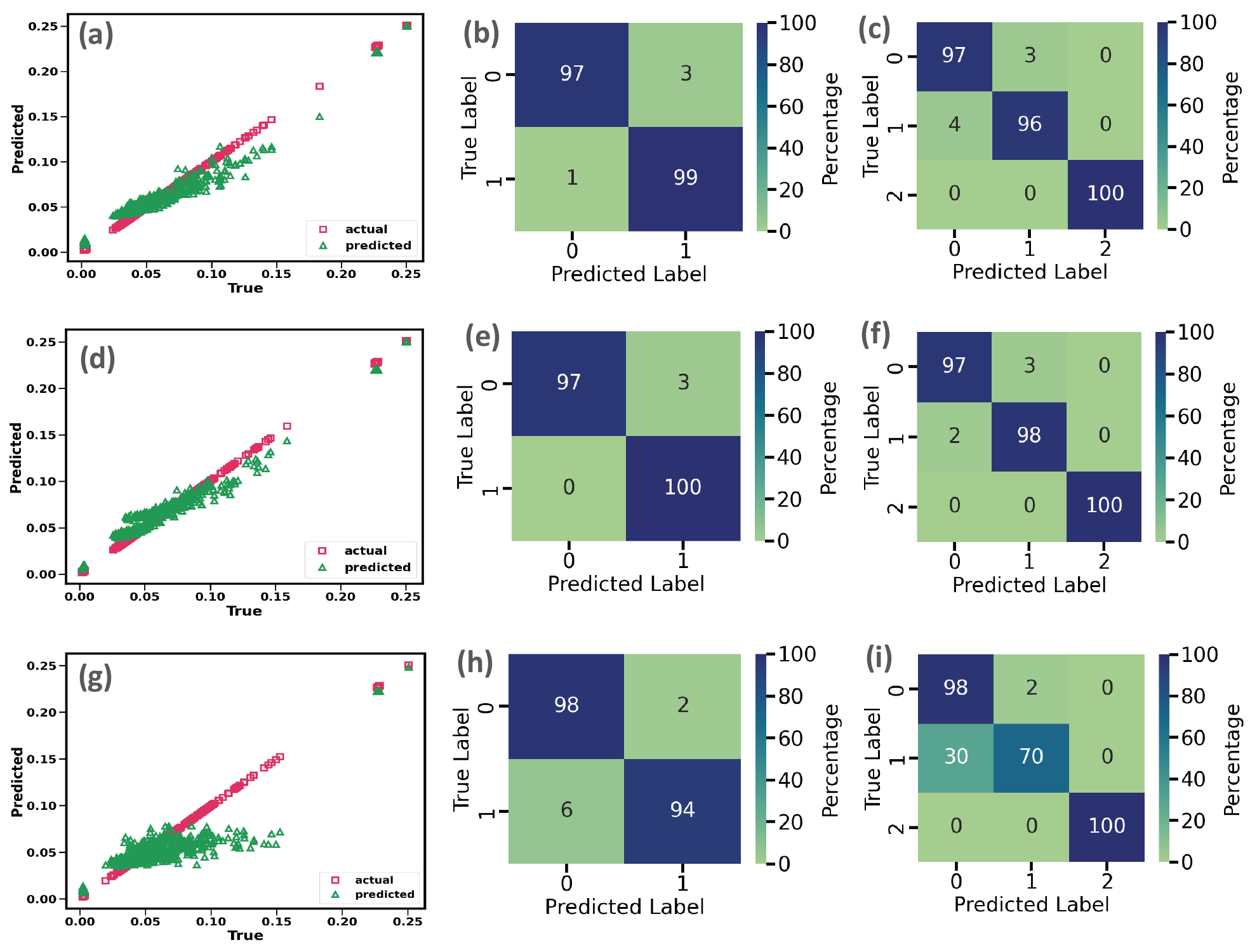}
\caption{Comparison of prediction performance using different subsets of node features. (a–c) show the results for the GAT model trained using only the top two features: betweenness and closeness centralities, demonstrating high accuracy in regression and classification. (d–f) show results using all five features, yielding similar classification accuracy but slightly better regression slope alignment with actual values. (g–i) display performance using only the bottom three features (degree, PageRank, and clustering coefficient), where classification between delocalized and localized is reasonable, but the model fails to accurately predict the weakly localized class.}
\label{fig:results_between_close}
\end{center}
\end{figure*}
Further, for the three-class classification in strong localized, weak localized, and delocalized states the GAT model trained with the closeness feature shows the highest accuracy (Fig. \ref{fig:cm_single_feature}(c)) and clustering coefficient shows the lowest (Fig. \ref{fig:cm_single_feature}(d)). The ranking of the features based on accuracy is as closeness ($\sim  95.3\%$), betweenness ($ \sim 94\%$), degree centrality ($\sim 68\%$), PageRank ($\sim 66\%$), and  clustering coefficient ($\sim 57\%$).

Combining the accuracy from binary and three-class classification shows that the top two important features are betweenness and closeness, and the cluster coefficient is the least important. To verify the impact of the top two features on the overall prediction accuracy, we have further trained a GAT model with these two features only. Fig. \ref{fig:results_between_close}(a-c) shows high prediction accuracy with these two features, which indicates that among all five features, these two features are more relevant and one can only choose these two features to design the solution framework. 
\begin{figure*}[tbh]
\begin{center}
\includegraphics[width = 7in, height = 3.6in]{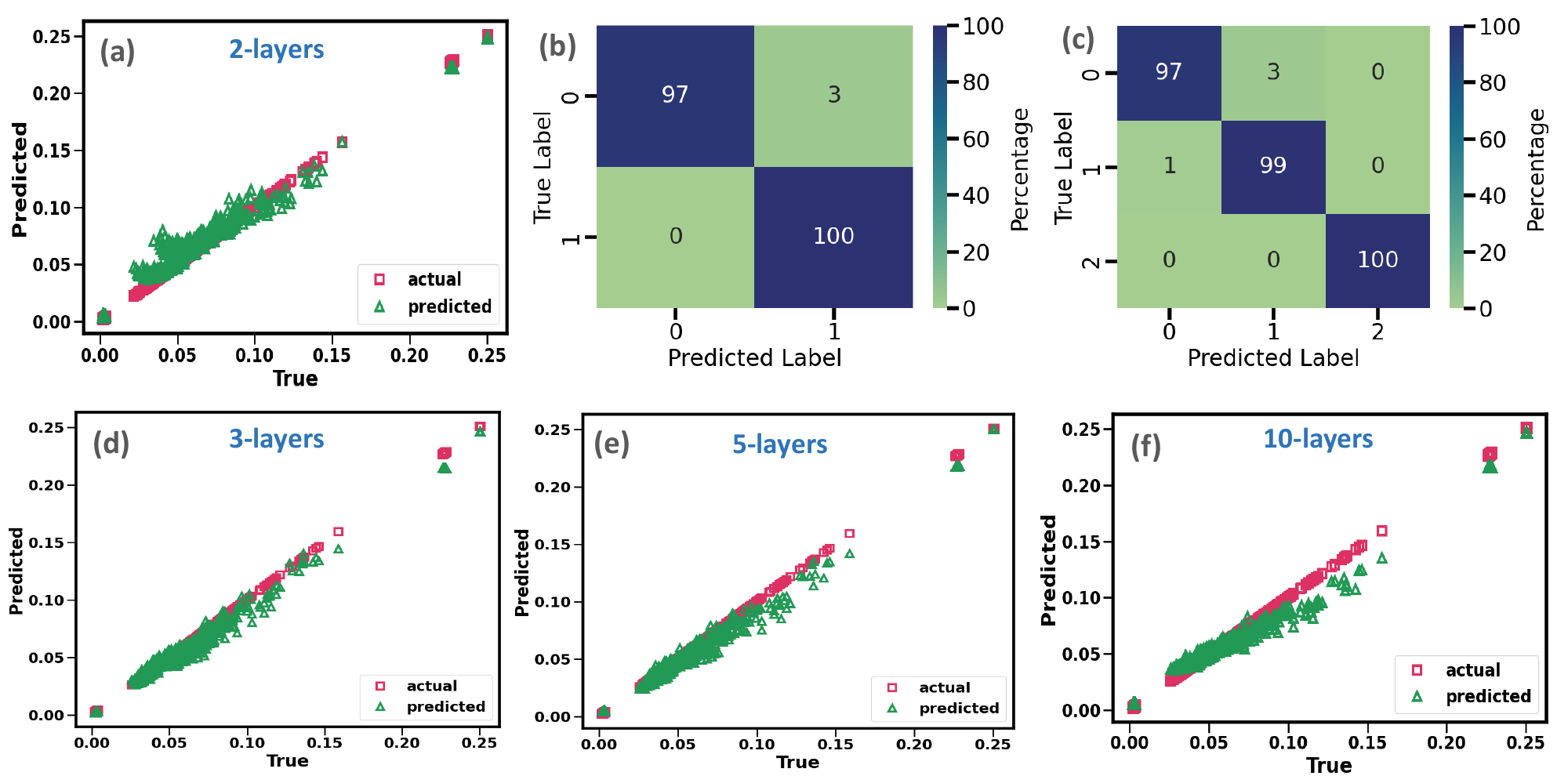}
\caption{Prediction performance of the GAT model after hyperparameter optimization using Optuna. (a) represent the predicted versus actual values of the IPR,  (b) confusion matrix of the binary classification and (c) represents the confusion matrix of delocalized, weakly localized, and strongly localized states, respectively. The optimized model uses $256$ hidden channels, $(4, 2)$ attention heads in the two GAT layers, dropout rate of $0.48$, and learning rate of $3 \times 10^{-3}$. (d-f) Illustrates the results for $3$, $5$, and $10$-layer GAT, respectively.}
\label{fig:results_optuna}
\end{center}
\end{figure*}
Further, we compared the accuracy of this experiment against the trained GAT model with all five features (Fig. \ref{fig:results_between_close}(d-f)). It shows similar accuracy to that obtained from the top two features. However, the classification accuracy may not increase much with all five features, but the slope of the regression line towards the actual values is slightly improved due to the use of all five features. Therefore, choosing all five features will help to enhance the solution with marginally improved accuracy. Finally, to understand the impact of the bottom three features (i.e., degree centrality, pagerank, and cluster coefficient), we have trained a GAT model with only these three features and observed that it hugely impacts the regression solution (Fig. \ref{fig:results_between_close}(g-i)) and consequently the three-class classification. The regression values for delocalized and strongly localized are predicted well; hence, the accuracy of binary classification is good; however, it poorly predicts the weakly localized, and as a result, the prediction accuracy of multi-classification is low. Note that the training data set comprised networks with sizes ranging from $200$ to $300$ nodes, with $1000$ samples per network type (cycle, path, wheel, star, ER, and SF), totaling $6000$ networks. The test data set included networks between $400$ and $500$ nodes, with $500$ samples per network type.

\textbf{Hyperparameter Optimization:}
We employed Optuna \cite{akiba2019optuna}, a hyperparameter optimization framework, to systematically fine-tune our GAT model. The optimization has been carried out based on both architectural and training-related hyperparameters. We designed two experimental setup (a) in the first, we fixed the number of layers to $2$ and optimized other hyperparameters (hidden channels, attention heads per GAT layer, dropout rate, and learning rate), (b) in the second, we varied the number of layers from $1$ to $6$, in addition to tuning other parameters.

For the first set-up (2 layers), the best performance is achieved with $256$ hidden channels, $(4, 2)$ attention heads, a dropout rate of $\sim0.48$, and a learning rate of $\sim 3 \times 10^{-3}$. The corresponding accuracy is shown in Fig. \ref{fig:results_optuna}(a-c). Notably, this model only marginally outperformed the baseline configuration with 64 hidden channels and $(4, 1)$ attention heads (Fig. \ref{fig:results_between_close} (d-f)). In the second set-up (with variable layers), the best configuration included $5$ layers, $256$ hidden channels, $(2, 1)$ attention heads, a dropout rate of $0.41$, and a learning rate of $\sim 2 \times 10^{-3}$. The results for this model are shown in Fig. \ref{fig:results_optuna} (e).

From the above, we observe that  increasing the number of layers significantly raises (nearly $\sim 5$ times, with the same training dataset) training time due to the larger number of trainable parameters, and also classification accuracy does not improve substantially with more layers (Figs. \ref{fig:results_between_close} (d) and \ref{fig:results_optuna} (a, d-f)). Hence, we prefer to use $2$ or $3$ layers to balance performance and efficiency. This reinforces our earlier observation (Fig. \ref{fig:results_between_close} (d)) that shallower GAT architectures can be efficient and effective for our work.

\subsection{Insights of Training Process}
To understand the explainability of our model, we provide mathematical insights into the training process via forward and backward propagation to predict the IPR value. Our derivation offers an understanding of the updation of weight matrices. We perform the analysis with a single GCN layer, a readout layer, and a linear layer for simplicity. However, our framework can easily be extended to more layers.

\vspace{2mm}
\noindent {\bf Forward Propagation:}  

\vspace{2mm}
\noindent GCN Layer: ${\bf H}^{(1,i)} = \sigma(\hat{\bf A}^{(i)} {\bf H}^{(0,i)} {\bf W})$

\vspace{2mm}
\noindent Readout Layer: $\bm{z}^{(i)} = \frac{1}{n_i}\sum_{j=1}^{n_i} \bm{h}^{(1,i)}_j$

\vspace{2mm}
\noindent Linear Layer: $\hat{y}^{(i)} = \bm{z}^{(i)} {\bf W}^{(\text{lin})} + b$   

\vspace{2mm}
\noindent Loss function: $\mathcal{L} = \frac{1}{N} \sum_{i=1}^N (y^{(i)} - \hat{y}^{(i)})^2$

In the above, $\hat{\bf A}^{(i)}$ is the normalized adjacency matrix, ${\bf H}^{(0, i)}$ is the initial feature matrix, and {\bf W} is the learnable weight matrix. Further, $\bm{h}^{(1,i)}_j=\sigma\biggl(\sum_{k=1}^{n_i} \hat{a}^{(i)}_{jk} \bm{h}^{(0,i)}_k {\bf W}\biggr)$ is the feature vector of node $j$ in graph $i$ and $j^{th}$ row of updated feature matrix ${\bf H}^{(1,i)}$ (Appendix \ref{GCN_math_details}, Example 1). Further, ${\bf W}^{(\text{lin})}$ and $b$ are the learnable weights of the linear layer. Finally, $y^{(i)}$ is the true scalar value for $\mathcal{G}_i$ and $\hat{y}^{(i)}$ is the predicted IPR value.

\vspace{2mm}
\noindent {\bf Backward Propagation:} 
To compute the gradients to update the weight matrices, we apply the chain rule to propagate the error from the output layer back through the network layers. We calculate the gradient of loss with respect to the output of the linear layer as
\begin{equation}
\frac{\partial \mathcal{L}}{\partial \hat{y}^{(i)}} = \frac{2}{N} (\hat{y}^{(i)} - y^{(i)})    
\end{equation}

\begin{figure*}[tbh]
\begin{center}
\includegraphics[width = 7in, height = 1.8in]{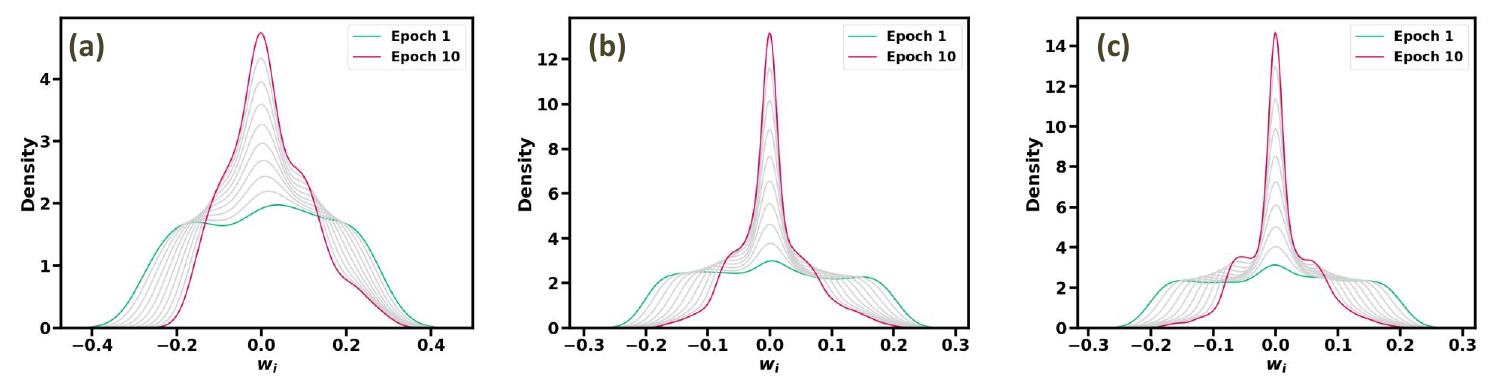}
\caption{Portray the distribution of weight matrices ($\mathbf{W}^{(0)},\mathbf{W}^{(1)},\mathbf{W}^{(2)}$) entries for the three GCN layers during the training process with cycle and star networks (Fig. \ref{loc_deloc_results_undirected}(a)). We  show the weights matrix entries for the first ten epochs. The gray color indicated weight matrices during epochs $2-9$.}
\label{weight_matrices}
\end{center}
\end{figure*}
We calculate the gradients for the linear layer. We know that each graph $i$ contributes to the overall loss $\mathcal{L}$. Therefore, we accumulate the gradient contributions from each graph when computing the gradient of the loss with respect to the weight matrix ${\bf W}^{(\text{lin})}$ (Appendix \ref{GCN_math_details}, Example 2). Thus, to obtain the gradient of the loss with respect to the weights \({\bf W}^{(\text{lin})}\), we apply the chain rule
\begin{equation}\nonumber 
\frac{\partial \mathcal{L}}{\partial {\bf W}^{(\text{lin})}} = \sum_{i=1}^N \biggl(\frac{\partial \mathcal{L}}{\partial \hat{y}^{(i)}} \cdot \frac{\partial \hat{y}^{(i)}}{\partial {\bf W}^{(\text{lin})}}\biggr) = \sum_{i=1}^N \frac{2}{N} (\hat{y}^{(i)} - y^{(i)}) z^{(i)}    
\end{equation}
where $\frac{\partial \hat{y}^{(i)}}{\partial {\bf W}^{(\text{lin})}} = z^{(i)}$. Similarly, we calculate the gradient with respect to $b$ and $z^{(i)}$ as 
\begin{equation}
\begin{split}
\frac{\partial \mathcal{L}}{\partial b} &= \sum_{i=1}^N \biggl(\frac{\partial \mathcal{L}}{\partial \hat{y}^{(i)}} \cdot \frac{\partial \hat{y}^{(i)}}{\partial b}\biggr) = \sum_{i=1}^N \frac{2}{N} (\hat{y}^{(i)} - y^{(i)}) \\
\frac{\partial \mathcal{L}}{\partial z^{(i)}} &= \frac{\partial \mathcal{L}}{\partial \hat{y}^{(i)}} \cdot \frac{\partial \hat{y}^{(i)}}{\partial \bm{z}^{(i)}} = \frac{2}{N} (\hat{y}^{(i)} - y^{(i)}) {\bf W}^{(\text{lin})}
\end{split}
\end{equation}
Now, we calculate the gradient for the Readout layer as
\begin{equation}\label{ap_f1}
\frac{\partial \mathcal{L}}{\partial \bm{h}^{(1,i)}_j} = \frac{\partial \mathcal{L}}{\partial \bm{z}^{(i)}} \cdot \frac{\partial \bm{z}^{(i)}}{\partial \bm{h}^{(1,i)}_j} = \frac{2}{N} (\hat{y}^{(i)} - y^{(i)}) {\bf W}^{(\text{lin})}\cdot \frac{1}{n_i}
\end{equation}
where $ \bm{z}^{(i)} = \frac{1}{n_i}\sum_{j=1}^{n_i} \bm{h}^{(1,i)}_j$ and thus $\frac{\partial \bm{z}^{(i)}}{\partial \bm{h}^{(1,i)}_j}=\frac{1}{n_i}$. 
Finally, we calculate the gradients for the GCN Layer. We have $N$ different graphs in our dataset, and each $\mathcal{G}_i$ has $n_i$ nodes. The total gradient with respect to ${\bf W}$ accumulates the contributions from all nodes in all graphs. Hence, we sum over all nodes in each graph and then over all graphs as 
\begin{equation}\label{ap_f}
\frac{\partial \mathcal{L}}{\partial {\bf W}} = \sum_{i=1}^N \sum_{j=1}^{n_i} \biggl(\frac{\partial \mathcal{L}}{\partial \bm{h}^{(1,i)}_j} \cdot \frac{\partial \bm{h}^{(1,i)}_j}{\partial {\bf W}}\biggr)    
\end{equation}
We know the layer output for the $i^{th}$ graph as ${\bf H}^{(1,i)} = \sigma(\hat{\bf A}^{(i)} {\bf H}^{(0,i)} {\bf W})$. Hence, for a single node $j$ in graph $i$, its node representation after the GCN layer is 
\begin{equation}\nonumber
\bm{h}^{(1,i)}_j = \sigma\left(\sum_{k=1}^{n_i} \hat{a}^{(i)}_{jk} \bm{h}^{(0,i)}_k {\bf W}\right)= \sigma(\bm{q}^{(i)}_j) 
\end{equation}
where $\bm{q}^{(i)}_j = \sum_{k=1}^{n_i} \hat{a}^{(i)}_{jk} \bm{h}^{(0,i)}_k {\bf W}$ and $\hat{a}^{(i)}_{jk}$ is the element in the $j^{th}$ row and $k^{th}$ column of the normalized adjacency matrix, representing the connection between node $j$ and node $k$ and $\bm{h}^{(0,i)}_k$ refers to the $k^{th}$ row of the input feature matrix \( {\bf H}^{(0,i)} \) of graph \( i \). To compute \(\frac{\partial \bm{h}^{(1,i)}_j}{\partial {\bf W}}\), we apply the chain rule as 
\begin{equation}\nonumber
\frac{\partial \bm{h}^{(1,i)}_j}{\partial {\bf W}} = \frac{\partial \bm{h}^{(1,i)}_j}{\partial \bm{q}^{(i)}_j} \cdot \frac{\partial \bm{q}^{(i)}_j}{\partial {\bf W}}
\end{equation}
We can calculate the partial derivative with respect to $\bm{q}^{(i)}_j$ as 
\begin{equation}\nonumber
\frac{\partial \bm{h}^{(1,i)}_j}{\partial \bm{q}^{(i)}_j} = \sigma'\left(\bm{q}^{(i)}_j\right)    
\end{equation}
where, $\sigma'(\bm{q}^{(i)}_j)$ is the derivative of $\sigma$. Now the partial derivative of $\bm{q}^{(i)}_j$ with respect to ${\bf W}$ as 
\begin{equation}\nonumber
\frac{\partial \bm{q}^{(i)}_j}{\partial {\bf W}} = \sum_{k=1}^{n_i} \hat{a}^{(i)}_{jk} \bm{h}^{(0,i)}_k    
\end{equation}
We can observe that \(\bm{q}^{(i)}_j\) is a linear combination of the rows of \({\bf H}^{(0,i)}\) weighted by \(\hat{\bf A}^{(i)}_j\). In matrix notation, we can write as
\begin{equation}\nonumber
\frac{\partial \bm{q}^{(i)}_j}{\partial {\bf W}} = \hat{\bm A}^{(i)}_j {\bf H}^{(0,i)}
\end{equation}
where $\hat{\bm A}^{(i)}_j$ is the $j^{th}$ row of ${\bf \hat{A}}^{(i)}$. Now, we combine the results of the chain rule and get
\begin{equation}\label{ap_f2}
\frac{\partial \bm{h}^{(1,i)}_j}{\partial {\bf W}} = \sigma'\left(\bm{q}^{(i)}_j\right) \cdot \hat{\bm A}^{(i)}_j {\bf H}^{(0,i)}
\end{equation}
In Eq. (\ref{ap_f}), we substitute Eqs. (\ref{ap_f1}) and (\ref{ap_f2}) and get
\begin{equation}
\begin{split}
\frac{\partial \mathcal{L}}{\partial {\bf W}} &= \frac{2}{N} \sum_{i=1}^N \sum_{j=1}^{n_i}  (\hat{y}^{(i)} - y^{(i)}) {\bf W}^{(\text{lin})}\cdot \frac{1}{n_i} \cdot \sigma'(\bm{q}^{(i)}_j)\\
&\cdot (\hat{\bm A}^{(i)}_j {\bf H}^{(0,i)})
\end{split}
\end{equation}
where $\bm{q}^{(i)}_j = \sum_{k=1}^{n_i} \hat{a}^{(i)}_{jk} \bm{h}^{(0,i)}_k {\bf W}$. Finally, the weight matrices are updated using gradient descent as

\begin{equation}
\begin{split}
{\bf W} &\leftarrow {\bf W} - \eta \frac{\partial \mathcal{L}}{\partial {\bf W}} \\
{\bf W}^{(\text{lin})} &\leftarrow {\bf W}^{(\text{lin})} - \eta \frac{\partial \mathcal{L}}{\partial {\bf W}^{(\text{lin})}}\\
b & \leftarrow b - \eta \frac{\partial \mathcal{L}}{\partial b}
\end{split}
\end{equation}
where $\eta$ is the learning rate. The above process is repeated iteratively: forward propagation $\rightarrow$ loss calculation $\rightarrow$ backward propagation $\rightarrow$ weight update until the model converges to an optimal set of weights that minimize the loss. For simplicity in backward propagation analysis, we use gradient-based optimization. However, all numerical results are reported using the Adam/AdamW optimization scheme. Recent research on backward propagation in GCN for node classification and link prediction can be found in \cite{hsiao2024derivation}.

We observe the weight matrices of different layers during the training time. The distribution of the weight matrices provides a visual representation of the weights learned during the training process (Fig. \ref{weight_matrices}). The magnitude of each weight indicates the importance of the corresponding feature. The higher absolute values in the weight matrices suggest that the feature significantly impacts the model's predictions. For an instance, we can observe that for different layers, weight matrix values are initially spread evenly around from zeros, but as time progresses, values become close to zeros (Fig. \ref{weight_matrices}). 

\section{Conclusion}
Using the graph neural network, we introduce a framework to predict the localized and delocalized states of complex unknown networks. We focus on leveraging the rich information embedded in network structures and extracting relevant features for graph regression. Specifically, a GCN model is employed to predict inverse participation ratio values for unknown networks and consequently identify localized or delocalized states. Our approach provides a graph neural network alternative to the traditional principal eigenvalue analysis \cite{loc2017optimized} for understanding the behavior of linear dynamical processes in real-world systems at steady state. This method offers near real-time insight into the structural properties of underlying networks. A key advantage of the proposed framework is its ability to train on small networks and generalize to larger networks, achieving an accuracy of nearly $ \sim 95 \%$ with test unseen model networks to understand delocalized or strongly localized states. This makes the model scale-invariant, with the computational cost of state prediction remaining consistent regardless of network size, apart from the cost of reading the network data. 

Our trained GNN framework (e.g., GCN and GAT) effectively identifies three different states in unseen test model network data. Moreover, our model accurately identifies the states in the weakly localized regions for the real-world data. However, distinguishing between delocalized weakly localized and strongly localized states associated with real-world graphs poses a significant challenge. It might be due to the imbalance of data points in different states and limited dataset availability. Further, alongside the IPR value prediction task, we have also incorporated Shannon entropy (SE) as a measure of localization, which is predicted using a GCN model (Appendix \ref{entropy_in_pev}). One can use Rényi entropy instead of SE, and this direction will be explored in the future. This work focuses solely on the principal eigenvector and provides a first proof of concept that GNN models can identify localized and delocalized states in linear dynamics on binary undirected graphs. The models can also be readily extended to study the localization behavior of other eigenvectors and to directed or weighted graphs~\cite{metz2021localization}.

\vspace{4mm}
\noindent \textbf{Code availability:} The code used in our experiments is available in the GitHub repository at \sloppy{\url{https://github.com/amitrezaiwf/Graph-Neural-Network-For-Steady-State-Identification}}, enabling reproducibility and facilitating future research built upon this work.

\begin{acknowledgments}
We thank the anonymous reviewer for their helpful comments, which have significantly improved the quality of this article. Priodyuti Pradhan is thankful to Sulthan Vishnu Sai (IIIT Raichur) for providing some of the model data generation codes, and is indebted to Anirban Dasgupta and Shubhajit Roy (Indian Institute of Technology Gandhinagar) for their insightful discussions. PP also acknowledges support from the Anusandhan National Research Foundation (ANRF), Govt. of India, under Grant No. TAR/2022/000657. Amit Reza is supported by the Netherlands Organization for Scientific Research (NWO). 
\end{acknowledgments}


\section*{References}
\bibliography{references}

\appendix
\renewcommand{\thefigure}{A\arabic{figure}}
\setcounter{figure}{0}
\renewcommand{\thetable}{A\arabic{table}}
\setcounter{table}{0}

\section{Linear Dynamics}\label{linear_dynamics_calculations}

We can write Eq. (\ref{Eq1:power_iteration}) in matrix form as
\begin{equation}
\frac{d\bm{x(t)}}{dt}={\bf M}\bm{x}(t)
\end{equation}
where {\bf M} is a transition matrix given by ${\bf M} =  \alpha{\bf I}+\beta{\bf A} $, where {\bf I} is the identity matrix. Note that {\bf M} and {\bf A} only differ by constant term. Hence,
\begin{equation}
\label{expo}
\bm{x}(t) = e^{{\bf M}t}\bm{x}(0)
\end{equation}
We consider ${\bf M}\in \mathbb{R}^{n \times n}$ is diagonalizable, ${\bf M}={\bf U \Lambda U^{-1}}$ and ${\bf U U^{-1}}={\bf I}$ where columns of {\bf U} are the eigenvectors ($\{\bm{u}_1^{\bf M},\bm{u}_2^{\bf M},\ldots,\bm{u}_n^{\bf M}\}$) of {\bf M} and having $n$ number of distinct eigenvalues $\{\lambda_1^{\bf M},\lambda_2^{\bf M},\ldots,\lambda_n^{\bf M}\}$ which are diagonally stored in $\Lambda$. We know $\bm{x}(0)$ is an arbitrary initial state. Thus, we can represent it as a linear combination of eigenvectors of {\bf M}, and therefore, we can write Eq. (\ref{expo}).
\begin{equation}
\label{perturbation_propagation}
\begin{split}
\bm{x}(t) & =  e^{{\bf M}t}[c_1(0)\bm{u}_1^{\bf M}+c_2(0)\bm{u}_2^{\bf M}+\ldots+c_n(0)\bm{u}_n^{\bf M}]\\
          & =  {\bf U}e^{{\bf \Lambda}t}{\bf U}^{-1}[c_1(0)\bm{u}_1^{\bf M}+c_2(0)\bm{u}_2^{\bf M}+\ldots+c_n(0)\bm{u}_n^{\bf M}]\\
          & =  {\bf U}e^{{\bf \Lambda}t}{\bf U^{-1}}{\bf U}\bm{c}(0)\\
          & =  {\bf U}e^{{\bf \Lambda}t}\bm{c}(0)\\
          & = c_1(0) e^{\lambda_1^{\bf M}t}\bm{u}_1^{\bf M} + c_2(0) e^{\lambda_2^{\bf M}t}\bm{u}_2^{\bf M} + \ldots+c_n(0) e^{\lambda_n^{\bf M}t}\bm{u}_n^{\bf M}\\
          &=\sum_{i=1}^{n}c_i(0)e^{\lambda_i^{\bf M}t}\bm{u}_i^{\bf M}
\end{split}
\end{equation}
such that where $\bm{c}(0)=(c_1(0),c_2(0),\ldots,c_n(0))^{T}$ and 
\begin{equation}
\begin{split}
e^{{\bf M}t} & =  {\bf I}+{\bf M}t+\frac{({\bf M}t)^2}{2}+\frac{({\bf M}t)^3}{3}+\ldots\\
             &=  {\bf I}+{\bf U \Lambda U^{-1}}t+\frac{({\bf U \Lambda U^{-1}}t{\bf U \Lambda U^{-1}}t)}{2!}+\\
             &\frac{({\bf U \Lambda U^{-1}}t{\bf U \Lambda U^{-1}}t{\bf U \Lambda U^{-1}}t)}{3!}+\ldots\\
             &=  {\bf U}\biggl[{\bf I}+{\bf \Lambda}t+\frac{({\bf  \Lambda} t)^2}{2!}+\frac{({\bf \Lambda} t)^3}{3!}+\ldots\biggr]{\bf U^{-1}}\\
             & ={\bf U}e^{{\bf \Lambda}t}{\bf U^{-1}}
\end{split}
\end{equation}
For $t \rightarrow \infty$, we can approximate Eq. (\ref{perturbation_propagation}) as  
\begin{equation}
\bm{x}^{*} \sim c_1(0)e^{\lambda_1^{\bf M}t}\bm{u}_1^{\bf M} \sim \bm{u}_1^{\bf M}
\end{equation}
Since the largest eigenvalue $\lambda_{1}$ dominates over the others, the PEV of the adjacency matrix will decide the steady state behavior of the system. 

\section{Mathematical Insights of Graph Convolution Neural Network}\label{GCN_math_details}
Deep Learning models, for example, Convolutional Neural Networks (CNN), require an input of a specific size and cannot handle graphs and other irregularly structured data \cite{graphclass2018}. Graph Convolution Networks (GCN) are exclusively designed to handle graph-structured data and are preferred over Convolutional Neural Networks (CNN) when dealing with non-Euclidean data. The GCN architecture draws on the same way as CNN but redefines it for the graph domain. Graphs can be considered a generalization of images, with each node representing a pixel connected to eight (or four) other pixels on either side. For images, the graph convolution layer also aims to capture neighborhood information for graph nodes. GCN can handle graphs of various sizes and shapes, which increases its applicability in diverse research domains.

The simplest GNN operators prescribed by Kipf et al. are called GCN  \cite{kipf2016semi}. The convolutional layers are used to obtain the aggregate information from a node's neighbors to update its feature representation. We consider the feature vector as $\bm{h}_i^{(l-1)}$ of node $i$ at layer $l-1$ and update the feature vector of node $i$ at layer $l$, as
\begin{equation}
\bm{h}_i^{(l)} = \sigma \left( \sum_{j \in \mathcal{N}(i) \cup \{i\}} \frac{1}{\sqrt{\tilde{d}_{i} \tilde{d}_{j}}} \bm{h}_j^{(l-1)} \mathbf{W}^{(l-1)} \right) \, ,
\end{equation} 
where new feature vector $\bm{h}_i^{(l)}$ for node $i$ has been created as an aggregation of feature vector $\bm{h}_i^{(l-1)}$ and the feature vectors of its neighbors $\bm{h}_j^{(l-1)}$ of the previous layer, each weighted by the corresponding entry in the normalized adjacency matrix ($\hat{{\bf A}}$), and then transformed by the weight matrix $\mathbf{W}^{(l-1)}$ and passed through the activation function $\sigma$. 
We use the ReLU activation function for our work.

The sum $\sum_{j \in \mathcal{N}(i) \cup \{i\}}$ aggregates the feature information from the neighboring nodes and the node itself where $\mathcal{N}(i)$ is the set of neighbors of node $i$. The normalization factor $1/\sqrt{\tilde{d}_{i} \tilde{d}_{j}}$ ensures that the feature vectors from neighbors are appropriately scaled based on the node degrees, preventing issues related to scale differences in higher vs. lower degree nodes where $\tilde{d}_{i}$ and $\tilde{d}_{j}$ being the normalized degrees of nodes $i$ and $j$, respectively \cite{GCNchaupham}. The weight matrix $\mathbf{W}^{(l-1)}$ transforms the aggregated feature vectors, allowing the GCN to learn meaningful representations. The activation function $\sigma$ introduces non-linearity, enabling the model to capture complex patterns.

\textbf{Single convolution layer representation:}
The operation on a single graph convolution layer can be defined using matrix notation as follows:
\begin{equation}\nonumber 
\begin{split}
{\bf H}^{(l)} &= \sigma \left( \hat{{\bf A}} {\bf H}^{(l-1)} {\bf W}^{(l-1)} \right)\\
&=\sigma \left(\tilde{\bf D}^{-\frac{1}{2}}\tilde{{\bf A}} \tilde{\bf D}^{-\frac{1}{2}} {\bf H}^{(l-1)} {\bf W}^{(l-1)}\right) 
\end{split}
\end{equation}
where $\mathbf{H}^{(l-1)}$ is the matrix of node features at layer $l-1$ where $l=1, 2, 3$, with \(\mathbf{H}^{(0)}\) being the input feature matrix. Here, $\tilde{{\bf A}} = {\bf A} + {\bf I}$ is the self-looped adjacency matrix by adding the identity matrix {\bf I} to the adjacency matrix {\bf A}. After that we do symmetric normalization by inverse square degree matrix with $\tilde{{\bf A}}$ and denoted as $\hat{{\bf A}} = \widetilde{\bf D}^{-\frac{1}{2}}\tilde{{\bf A}} \widetilde{\bf D}^{-\frac{1}{2}}$, where ${\bf D} \in \mathbb{R}^{n\times n}$ is the diagonal degree matrix of ${\bf A}$ with $\tilde{D}_{ii} = \sum_{j = 1}^{n} \tilde{a}_{ij}$. Here, ${\bf W}^{(l)} \in \mathbb{R}^{F_{\text{in}} \times F_{\text{out}}}$ is a trainable weight matrix of layer $l$. A linear feature transformation is applied to the node feature matrix by ${\bf HW}$, mapping the $F_{\text{in}}$ feature channels to $F_{\text{out}}$ channels in the next layer. The weights of ${\bf W}$ are shared among all vertices. We use the Glorot (Xavier) initialization that initializes the weights by drawing from a distribution with zero mean and a specific variance \cite{glorot2010understanding}. It helps maintain the variance of the activations and gradients through the layers for a weight matrix \(\mathbf{W}\) 
\begin{equation}
\mathbf{W} \sim \mathcal{U} \left( -\sqrt{\frac{6}{F_{\text{in}} + F_{\text{out}}}}, \sqrt{\frac{6}{F_{\text{in}} + F_{\text{out}}}} \right)    
\end{equation}
where, $\mathcal{U}$ denotes the uniform distribution. For the GCN layer implementation, we use GCNConv from the PyTorch Geometric library where adjacency matrix ${\bf A}^{(i)}$ is transformed to the edge lists for fast processing  \cite{FeyLenssen2019}.

\subsubsection{Example 1}\label{eg_1}
For instance, we consider matrices
\[
{\bf A} =
\begin{bmatrix}
1 & 2 \\
3 & 4
\end{bmatrix}, \quad
{\bf H} =
\begin{bmatrix}
1 & 0 & 2 \\
-1 & 3 & 1
\end{bmatrix}, \quad
{\bf W} =
\begin{bmatrix}
1 & 2 \\
0 & 1 \\
-1 & 0
\end{bmatrix}
\]
We compute
\[
{\bf D = A H W}
\]
as
\[
{\bf E = A H} =
\begin{bmatrix}
1 & 2 \\
3 & 4
\end{bmatrix}
\begin{bmatrix}
1 & 0 & 2 \\
-1 & 3 & 1
\end{bmatrix}
 =
\begin{bmatrix}
-1 & 6 & 4 \\
-1 & 12 & 10
\end{bmatrix}
\]

Now we compute 
\[
 {\bf D = E W}=
\begin{bmatrix}
-1 & 6 & 4 \\
-1 & 12 & 10
\end{bmatrix}
\begin{bmatrix}
1 & 2 \\
0 & 1 \\
-1 & 0
\end{bmatrix}
 =
\begin{bmatrix}
-5 & 4 \\
-11 & 10
\end{bmatrix}
\]

Now, we express the \( j \)th row of \( {\bf D} \) as
\[
\bm{d}_j = \sum_{k=1}^{n} A_{jk} \bm{h}_k {\bf W}
\]
For \( j = 1 \) (first row of \( {\bf D} \))
\[
\bm{d}_1 = A_{11} \bm{h}_1 {\bf W} + A_{12} \bm{h}_2 {\bf W}
 = (1 \bm{h}_1 + 2 \bm{h}_2) {\bf W}
\]

We know rows of ${\bf H}$ as 
\[
\bm{h}_1 =
\begin{bmatrix}
1 & 0 & 2
\end{bmatrix}, \quad
\bm{h}_2 =
\begin{bmatrix}
-1 & 3 & 1
\end{bmatrix}
\]
Hence,
\[
1 \bm{h}_1 + 2 \bm{h}_2 = 
\begin{bmatrix}
1 & 0 & 2
\end{bmatrix}
+ 2 \times
\begin{bmatrix}
-1 & 3 & 1
\end{bmatrix}
=
\begin{bmatrix}
-1 & 6 & 4
\end{bmatrix}
\]

Now, multiplying with \( {\bf W} \) we get
\[
\bm{d}_1 = (-1,6,4) \times
\begin{bmatrix}
1 & 2 \\
0 & 1 \\
-1 & 0
\end{bmatrix}
=
\begin{bmatrix}
-5 & 4
\end{bmatrix}
\]

Similarly, for \( j = 2 \), we get
\[
\bm{d}_2 = (-1,12,10) \times {\bf W} = \begin{bmatrix} -11 & 10 \end{bmatrix}
\]

\subsubsection{Example 2}\label{eg_2}

Let's consider an example with three graphs, each having a corresponding \(z^{(i)}\) and \(y^{(i)}\) as $z^{(1)}, z^{(2)}, z^{(3)}$ are the outputs from the readout layer and $y^{(1)}, y^{(2)}, y^{(3)}$ are the true scalar values. If we denote the linear layer weight as \({\bf W}^{(\text{lin})}\), thus we can compute the predictions as 
\begin{equation}
 \begin{split}
 \hat{y}^{(1)} &= z^{(1)} {\bf W}^{(\text{lin})} + b \\
 \hat{y}^{(2)} &= z^{(2)} {\bf W}^{(\text{lin})} + b\\
 \hat{y}^{(3)} &= z^{(3)} {\bf W}^{(\text{lin})} + b
 \end{split}
\end{equation}
Now, we can compute gradients for each graph:
\begin{equation}
 \begin{split}
 \frac{\partial \mathcal{L}}{\partial \hat{y}^{(1)}} &= \frac{2}{3} (\hat{y}^{(1)} - y^{(1)}) \\
 \frac{\partial \mathcal{L}}{\partial \hat{y}^{(2)}} &= \frac{2}{3} (\hat{y}^{(2)} - y^{(2)})\\
 \frac{\partial \mathcal{L}}{\partial \hat{y}^{(3)}} &= \frac{2}{3} (\hat{y}^{(3)} - y^{(3)})
 \end{split}
\end{equation}
Hence, aggregate gradients for ${\bf W}^{(\text{lin})}$ as
\begin{equation}\nonumber
 \begin{split}
 \frac{\partial \mathcal{L}}{\partial {\bf W}^{(\text{lin})}} &= \left( \frac{2}{3} (\hat{y}^{(1)} - y^{(1)}) \right) z^{(1)} + \left( \frac{2}{3} (\hat{y}^{(2)} - y^{(2)}) \right) z^{(2)} \\
 &+ \left( \frac{2}{3} (\hat{y}^{(3)} - y^{(3)}) \right) z^{(3)}
 \end{split}
\end{equation}

\section{Mathematical Insights of Graph Attention Network}\label{GAT_math_details}

Graph Attention Networks (GATs) are an extension of Graph Convolutional Networks (GCNs) that introduce attention mechanisms to improve message passing in graph neural networks \cite{velivckovic2017graph}. The key advantage of GAT is that it assigns different importance (attention) to different neighbors, making it more flexible and powerful than traditional GCNs, which use fixed aggregation weights.

In a standard GCN, node embeddings are updated by aggregating information from neighboring nodes using fixed weights derived from the adjacency matrix. In a GAT, an attention mechanism is used to dynamically compute different weights for each neighbor, allowing the network to focus more on important neighbors. In GAT, each node feature vector ($\bm{h}_i$) is transformed into a higher-dimensional representation using a learnable weight matrix ${\bf W}$ as
\begin{equation}\nonumber
\bm{h}_i' = {\bf W} \bm{h}_i  
\end{equation}
where \( {\bf W} \in \mathbb{R}^{F' \times F} \) is a learnable weight matrix, and \( F' \) is the new feature dimension. Further, for each edge \( (i, j) \in E \), compute the attention score \( e_{ij} \), which measures the importance of node \( j \) 's features to node \( i \). The attention score is calculated as
\[
e_{ij} = \text{LeakyReLU}(\mathbf{a}^T [ {\bf W} \bm{h}_i \| {\bf W} \bm{h}_j ])
\]
where \( \mathbf{a} \in \mathbb{R}^{2F'} \) is a learnable weight vector, \( \| \) denotes concatenation, and LeakyReLU is used as a non-linear activation function. Finally, the attention scores are normalized across all neighbors using the softmax function.
\[
\alpha_{ij} = \frac{\exp(e_{ij})}{\sum_{k \in \mathcal{N}(i)} \exp(e_{ik})}
\]
where \( \mathcal{N}(i) \) denotes the neighbors of node \( i \). The softmax ensures that the attention weights sum to $1$ for each node. Each node aggregates its neighbors' transformed features using the learned attention coefficients.
\[
\bm{h}_i' = \sigma \left( \sum_{j \in \mathcal{N}(i)} \alpha_{ij} {\bf W} \bm{h}_j \right)
\]
where \( \sigma \) is a non-linearity (e.g., ReLU). To improve stability, GAT often uses multi-head attention, where multiple attention mechanisms run in parallel, and their outputs are averaged.
\[
\bm{h}_i' = \sigma \left( \frac{1}{K} \sum_{k=1}^{K} \sum_{j \in \mathcal{N}(i)} \alpha_{ij}^{(k)} {\bf W}^{(k)} \mathbf{h}_j \right)
\]
where $K$ is the number of attention heads, ${\bf W}^{(k)}$ and $\alpha_{ij}^{(k)}$ are the weight matrix and attention coefficients of the $k^{th}$ attention head. For our graph-level IPR value regression task, we use two layers of GAT with four heads for the first layer and one head in the second layer, respectively. For the GAT layer implementation, we use GATConv from the PyTorch Geometric library \cite{FeyLenssen2019}.

\begin{figure*}[tbh]
\begin{center}
\includegraphics[width = 6in, height = 3.2in]{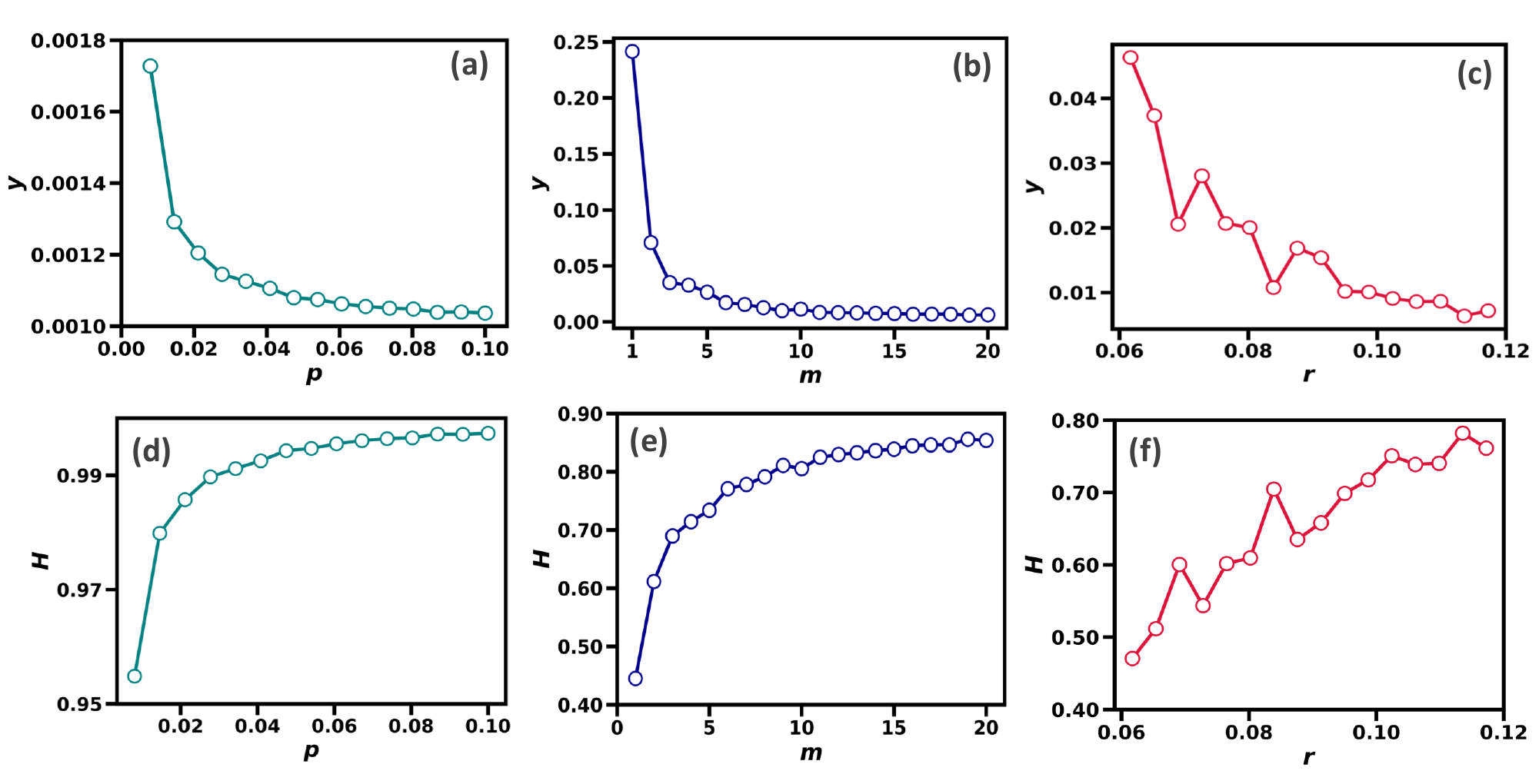}
\caption{IPR and entropy measures across different model networks. (a) In ER random graphs, increasing the connection probability $p$ leads to more edges and a decrease in IPR values, 
(b) In scale-free networks, higher attachment parameter $m$ results in lower IPR due to increased average degree. 
(c) A similar trend is seen in RGGs as the connection radius grows. 
(d–f) In contrast, the entropy measure (Eq.~\ref{shannon_entropy}) increases with connectivity, showing an inverse relationship to IPR. Here, we consider network size, $n=1000$.}
\label{IPR_Entropy_linear_dynamics}
\end{center}
\end{figure*}

\section{Complex Networks}\label{complex_networks}
We prepare the datasets for our experiments using several model networks (cycle, path, star, wheel, ER, and scale-free networks) and their associated IPR values. A few models (ER and scale-free networks) are randomly generated. 

\subsubsection{ER random network}

The Erdős–Rényi (ER) random network is denoted by $\mathcal{G}^{ER}(n,p)$, where $n$ is the number of nodes and $p$ is the edge probability \cite{posfai2016network}. Each edge exists independently of others. Starting with $n$ nodes, edges are connected with probability $p = \langle k \rangle / n$, where $\langle k \rangle$ is the mean degree, resulting in a Binomial degree distribution \cite{posfai2016network}. We generate ER random graphs of size $n$, starting with the smallest $p$ values, ensuring a connected graph, and incrementally increasing $p$ values to study the IPR of PEV. The connectivity threshold for $G(n, p)$ is $p_{c} \approx \frac{\ln n}{n}$. We start from $p > p_{c}$ and increase it gradually. It implies that low $p$ (sparse graphs) leads to a slightly higher IPR. On the other hand, high $p$ (i.e., for dense graphs) leads to eigenvector spread out, low IPR (Fig. \ref{IPR_Entropy_linear_dynamics} (a)). We can observe that from low to high $p$, there are tiny changes in the IPR values that lead to a delocalized state.

\subsubsection{Scale-free random network}
Scale-free (SF) networks ($\mathcal{G}^{SF}$) generated via the Barabási–Albert (BA) model exhibit a power-law degree distribution \cite{posfai2016network}. The construction begins with a small connected graph of $m_{0}$ nodes. Each step introduces a new node and links to $m$ existing nodes via preferential attachment. To study how connectivity influences eigenvector localization, we vary the number of edges added per node ($m$), which controls network density. As $m$ increases, the network becomes denser, and the average degree scales approximately as $\langle k \rangle \approx 2m$. We compute the IPR of PEV of the adjacency matrix for networks generated with different $m$. We observe (Fig. \ref{IPR_Entropy_linear_dynamics} (b)) that the smaller values of $m$ lead to localized eigenvectors (i.e., higher IPR values), whereas the larger values induce delocalization (i.e., smaller IPR values), indicating a broader participation of nodes in the leading eigenvector. This behaviour highlights the critical role of degree heterogeneity and connectivity in shaping localization phenomena in complex networks.

\subsubsection{Random Geometric Graph}
The Random Geometric Graph (RGG) model represents spatial networks by embedding nodes in a geometric space and connecting them based on proximity \cite{dall2002random}. We construct an RGG, $G(n, r)$ consists of $n$ nodes placed uniformly at random in a unit square $[0, 1]^{2}$, where any two nodes $v_{i}, v_{j}$ are connected if their Euclidean distance satisfies $\|\bm{x}_{i} - \bm{x}_{j} \| \leq r$, where $r$ is connection radius (i.e., a threshold) and each node $v_{i}$ has a random position $\bm{x}_{i} \in \mathbb{R}^{2}$. Note that instead of Euclidean distance metric, one can use other distance matrices to construct an RGG. The adjacency matrix ${\bf A}$ is defined as
$$
a_{ij} = 
\begin{cases}
1, & \text{if } i \neq j \text{ and } \|\bm{x}_i - \bm{x}_j\| \leq r, \\
0, & \text{otherwise.}
\end{cases}
$$
The network becomes connected asymptotically when the radius scales as $r(n) \gtrsim \sqrt{\ln n / (\pi n)}$. RGGs exhibit high clustering and a narrow degree distribution due to spatial constraints, with most nodes having similar degrees. To study eigenvector localization, we generate RGGs with increasing $r$, starting from the theoretical minimum $r_{\text{min}} \approx \sqrt{\ln n / (\pi n)}$ for connectivity. For each $r$, we compute IPR of the PEV, discarding disconnected realizations. As $r$ grows, the IPR decreases, reflecting delocalization (Fig. \ref{IPR_Entropy_linear_dynamics}(c)).

\begin{figure*}[tbh]
\begin{center}
\includegraphics[width = 6in, height = 4.3in]{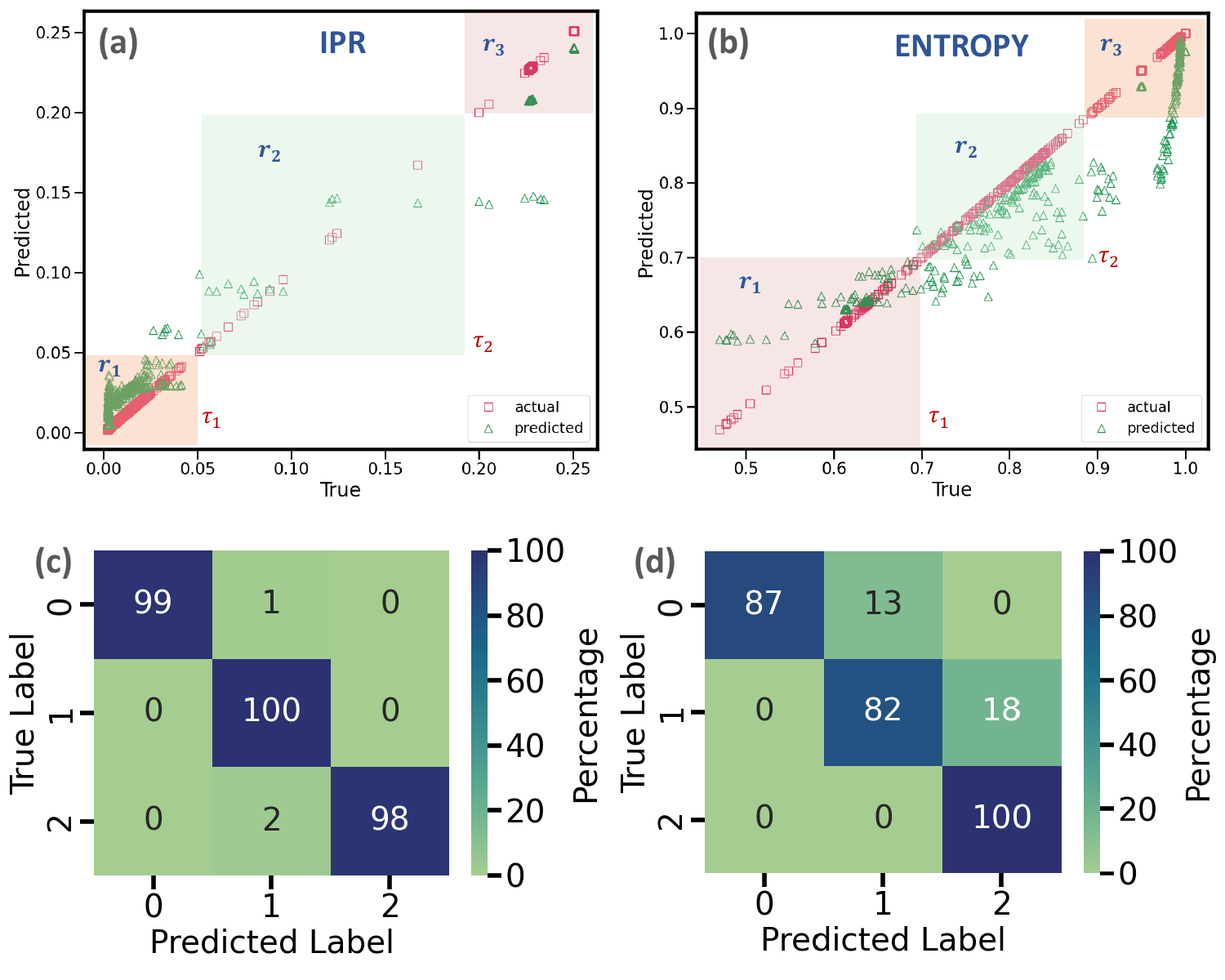}
\caption{GCN model predictions on network datasets. Test performance of the GCN model trained on networks generated as in Fig.~\ref{IPR_Entropy_linear_dynamics}, including star, cycle, path, wheel, ER, SF, and RGG topologies. (a, c) show true versus predicted values for the inverse participation ratio (IPR), while (b, d) illustrate results for the normalized Shannon entropy. The GCN captures trends for extreme values but shows reduced accuracy across intermediate regimes. For the IPR measure, the classes are denoted as delocalized ($r_1 \mapsto 0$), weakly localized ($r_2 \mapsto 1$), and strongly localized ($r_3 \mapsto 2$). However, for the entropy measure, the regions are reversed, delocalized ($r_3 \mapsto 0$), weakly localized ($r_2 \mapsto 1$), and strongly localized ($r_1 \mapsto 2$).
}
\label{GCN_IPR_ENTROPY}
\end{center}
\end{figure*}

Finally, we incorporate different generated model networks as discussed above. The datasets contain cycle, star, wheel, path, ER random, SF, and random geometric graphs, where the traning data sets size is $1400$, where each of type is $200$ and each graph size varies from $200-300$ and test datasets contains $700$ where each of type $100$ and size varies from $400-500$. We train a GCN model to distinguish strongly localized, weakly localized, and delocalized states and observe that the model predicts the IPR values and the threshold function identifies the classes very well for the IPR values (Fig. \ref{GCN_IPR_ENTROPY}(a, c)).

\begin{figure*}[tbh!]
\begin{center}
\includegraphics[width = 7in, height = 5in]{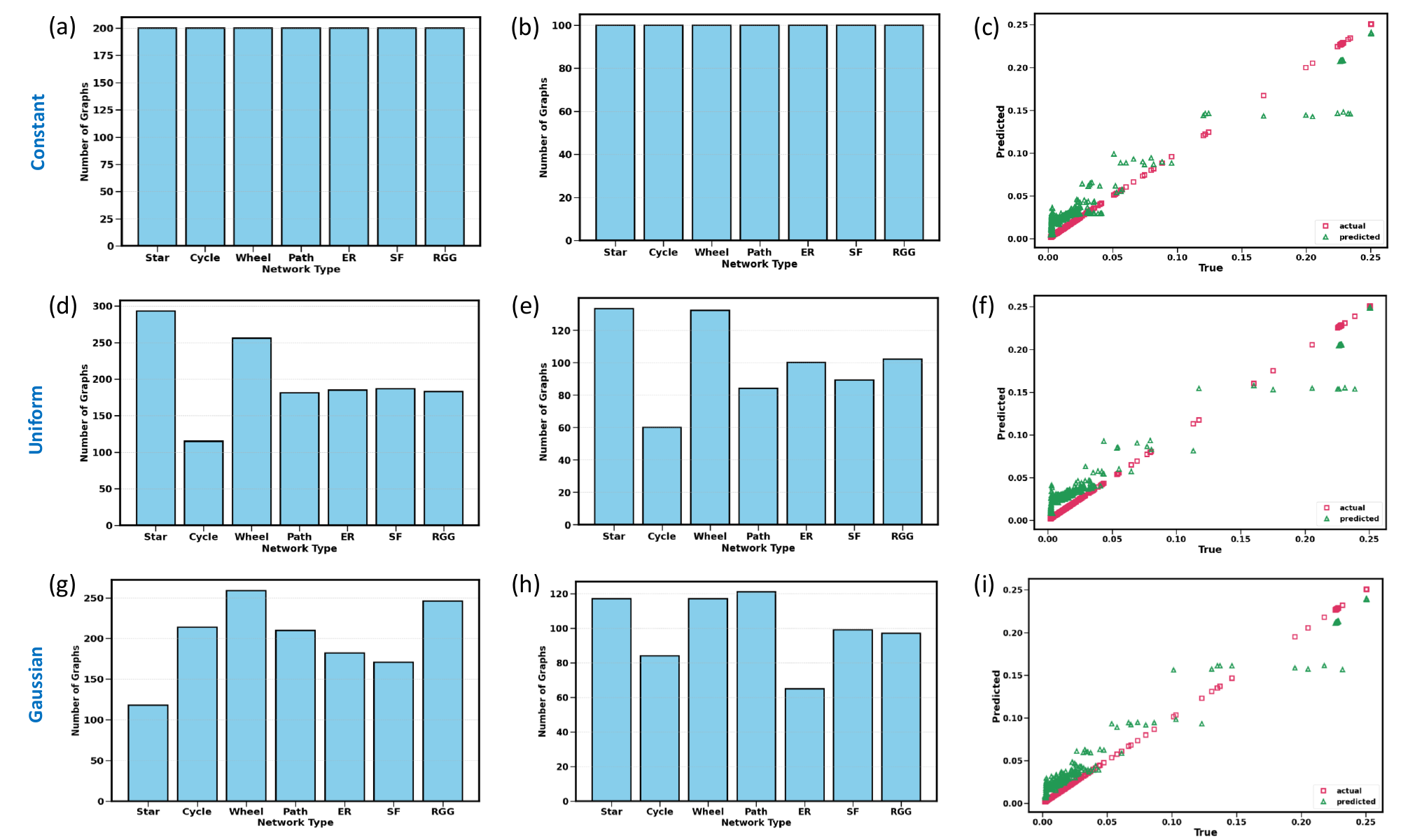}
\caption{Distribution of different network types in the training and test datasets. Each of the seven model network types (Star, Cycle, Wheel, Path, ER, SF, RGG) is represented with (a-c) equal frequency in both datasets to ensure balanced training and evaluation (d-f) from uniform distribution, and finally (g-i) portrayed from the Gaussian distribution.}
\label{train_test_data_dist}
\end{center}
\end{figure*}

For our experiment, we consider three different distribution of network types in the training and test datasets (Fig.~\ref{train_test_data_dist}). In the primary setup, we considered equal-sized subsets for each network type to eliminate bias due to class imbalance (Fig.~\ref{train_test_data_dist}(a-c)). We have also conducted a follow-up experiment with subset sizes drawn from uniform and Gaussian distributions ((Fig.~\ref{train_test_data_dist})(d-i)). The results, shown in Fig.~\ref{train_test_data_dist} indicate that the model remains robust across different data distributions, demonstrating a reasonable level of generalization even when the training and testing composition is skewed.

\section{Random features vs. Network features}\label{random_vs_network_features}
We conducted a series of experiments with varying node feature types to understand the role of node features in the learning process of GNN. We begin with zero features for each node, followed by constant features, and then experimented with random and network-specific features. The choice of zero-feature setting is inspired by the null solution scheme, which means our expectation is that the GNN model fails to learn in this specific case. For the zero-feature setting, where each node has input $h_{i} = (0, 0, \ldots, 0)$, the initial representation ${\bf H}^{(0)}$ becomes the zero matrix, and we set the bias term in the linear layer to zero. As a result, subsequent layers ${\bf H}^{(l)}$ in Eq.~(\ref{gcn_layers}) remain zero due to linearity and ReLU activation, leading to no learning. 

\begin{figure*}[tbh!]
\begin{center}
\includegraphics[width = 7in, height = 4.8in]{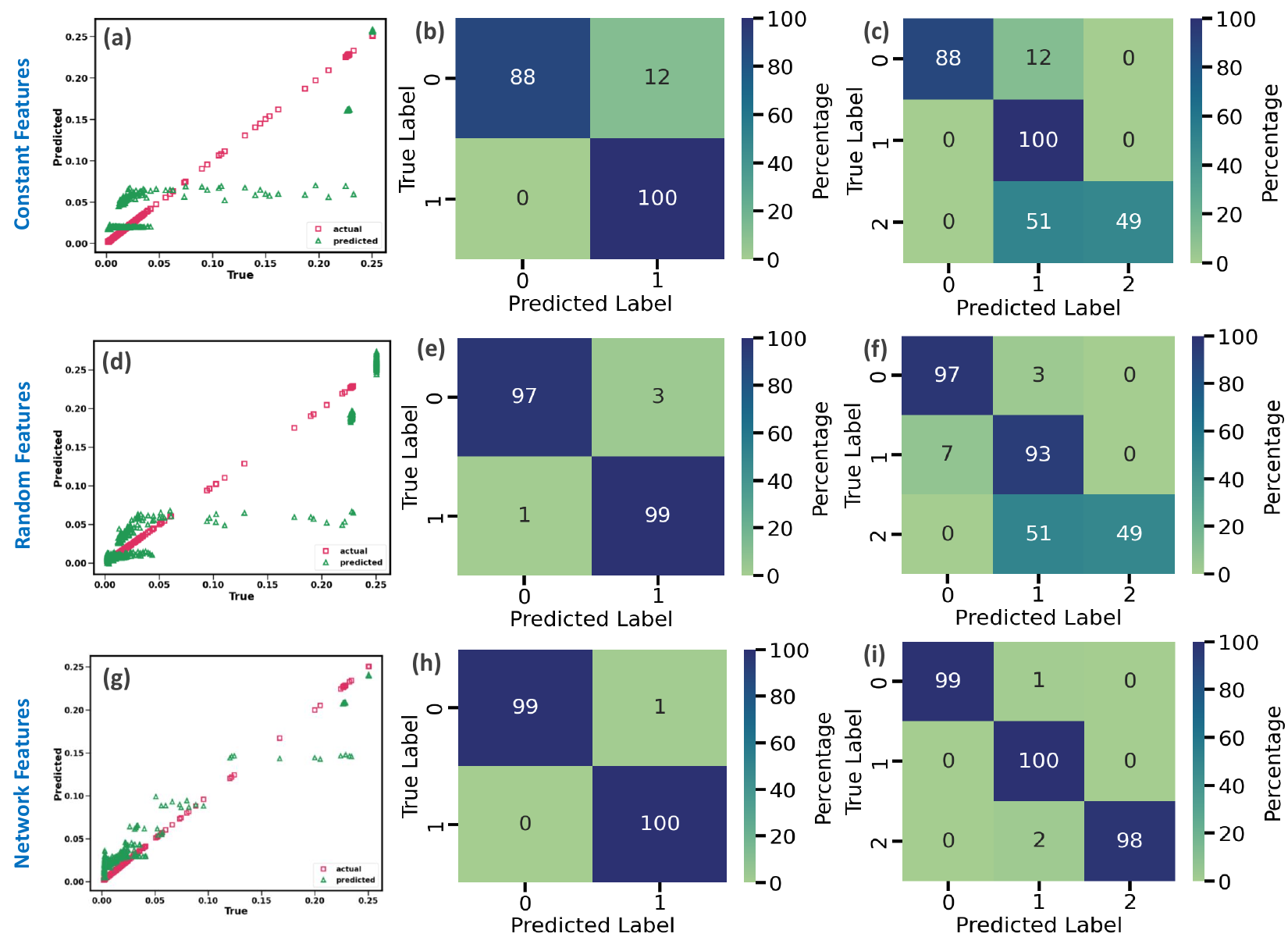}
\caption{Constant vs. Random vs. Network features. Our training and test datasets consist of model networks (Star, Cycle, Wheel, Path, ER, SF, and RGG). The training set contains $1,400$ graphs ($200$ from each topology) with sizes ranging from $200$ to $300$ nodes, while the test set comprises $700$ graphs ($100$ from each topology) with sizes between $400$ and $500$ nodes. We train the model for $100$ epochs using the AdamW optimizer with a learning rate of $10^{-3}$ and a weight decay of $5 \times 10^{-4}$. Training converges with the loss approaching zero.}
\label{constant_random_vs_network_features}
\end{center}
\end{figure*}

With constant-feature setting, each node is initialized with constant input, such as $h_{i} = (1, 1, \ldots, 1)$. This eliminates node-level distinguishability, allowing us to understand the impact of graph topology alone. That implies that any predictive ability of the GNN must arise solely from message passing over the graph structure, providing a structural baseline. We have trained a GCN model with zero and constant features, and as expected, obtained no learning with zero feature setting. With constant feature setting, GCN reaches a limited performance (Fig. \ref{constant_random_vs_network_features}(a-c)). Furthermore, we have trained a GCN model with random binary node features (i.e., $0$ or $1$), and network-specific features such as degree, closeness, betweenness centralities, PageRank, and clustering coefficients. Fig.~\ref{constant_random_vs_network_features}(d-i) shows that the GCN achieves significantly better accuracy when network-specific features are used than random features.

\section{Entropy for Localization Analysis in Complex Networks}\label{entropy_in_pev}

\textbf{Shannon Entropy.} The Shannon entropy is a statistic to measure the spread of a wavefunction or probability distribution in space \cite{anand2009entropy}. This measure is widely used in quantum mechanics, statistical physics, and information theory to characterize localization-delocalization of transitions. 
Let $\bm{p} = (p_{1}, p_{2}, \ldots, p_{n})$ be a probability vector such that $\sum_{i=1}^n p_{i} = 1$, hence the Shannon entropy can be defined as 
\begin{equation}\label{shannon_entropy}
H (\bm{p})= -\sum_{i=1}^{n} p_i \ln(p_i) \, ,  
\end{equation}
where \(n\) is the number of nodes, $p_{i} = |u_{i}|^2$ is the squared magnitudes of the components of the eigenvector. It gives a measure of uncertainty or spread of the eigenvector over nodes. For example, fully localized eigenvector, $\bm{u} = (1, 0, 0, \ldots, 0)$, has $H = -1 \cdot \ln(1) = 0$ leads to minimum entropy value. For fully delocalized eigenvector, $\bm{u} = (\frac{1}{\sqrt{n}}, \frac{1}{\sqrt{n}},\ldots, \frac{1}{\sqrt{n}})$, yields  
$H = -\sum_{i=1}^n \frac{1}{n} \ln\left(\frac{1}{n}\right) = \ln(n)$, which is maximum entropy value.  Hence, for a network of size \( n \), when the Shannon entropy \(H\) of a state lies strictly between the fully localized and fully delocalized (\(0 \ll H \ll \ln n\)), it indicates partial localization or weak localization. So entropy increases with delocalization,  and is inversely related to IPR (Table \ref{loc_measure_relation}). To facilitate comparison across networks of different sizes, we use the normalized form
\begin{equation}
H^{\text{norm}} = \frac{H}{\ln n} \, ,
\end{equation}
where $n$ is the number of nodes in the network. As we consider the network is connected, we never get entropy ($H \equiv H^{norm}$) to be zero and thus $0 < H \le 1$. Now, to identify the states ($\bm{p} \equiv \bm{x}^{*}$) belong to which category of dynamical behavior for linear dynamics, we can modify the threshold scheme for identifying entropy values  lies in the range $(0, 1]$.
\begin{table*}[tbh]\label{loc_measure_relation}
\centering
\begin{tabular}{@{}lll@{}}
\toprule
\textbf{Measure} & \textbf{Formula} & \textbf{Localization Implication} \\
\midrule
Inverse Participation Ratio (IPR) & $\sum_i p_i^2$     & $1 \Rightarrow$ localized,\quad $1/n \Rightarrow$ delocalized \\
Shannon Entropy  & $-\sum_i p_i \ln p_i$           & $0 \Rightarrow$ localized,\quad $\ln n \Rightarrow$ delocalized \\
Rényi Entropy ($q=2$) & $-\ln \sum_i p_i^2$             & $0 \Rightarrow$ localized,\quad $\ln n \Rightarrow$ delocalized \\
\bottomrule
\end{tabular}
\caption{Summary of entropy-based localization measures in complex networks.}
\end{table*}

\vspace{2mm}
We follow a threshold scheme like IPR (Eq. \ref{threshold_scheme}) for dividing states into strongly, weakly localized, and delocalized regimes. We define two thresholds, $\tau_{1}$ and $\tau_{2}$, such that $0 < \tau_{1} < \tau_{2} \le 1$. An additional parameter $\epsilon$ ($\epsilon > 0$) defines some flexibility around the thresholds.

\vspace{0.2cm}
\noindent {{\em \textbf{Strongly localized region}} ($r_{1}$): 
We consider strongly localized state if entropy values are significantly below the first threshold, including an $\epsilon$-width around $\tau_{1}$, i.e., 
\begin{equation}\nonumber
r_{1} = \{H \in (0, 1] \mid H \leq \tau_{1} - \epsilon \}    
\end{equation}
\noindent {{\em \textbf{Weakly localized region}} ($r_{2}$):} 
We consider weakly localized state if entropy values around and between the two thresholds, including $\epsilon$-width around $\tau_{1}$ and $\tau_{2}$, i.e., 
\begin{equation}\nonumber
r_{2} = \{H \in (0, 1] \mid \tau_{1} - \epsilon < H < \tau_{2} + \epsilon \}
\end{equation}
\noindent {{\em \textbf{Delocalized region}} ($r_{3}$):}
If entropy values are significantly above the second threshold, we consider delocalized states, including an $\epsilon$-width around $\tau_{2}$, i.e.,
\begin{equation}\nonumber
r_{3} = \{H \in (0, 1] \mid H \geq \tau_{2} + \epsilon \}
\end{equation}  

For instance, we consider a set of threshold values as $\tau_{1} = 0.7$, $\tau_{2} = 0.9$, $\epsilon = 1e-6$. For example, in a star graph, PEV  $\bm{u}^{\mathcal{S}}=\biggl(\frac{1}{\sqrt{2}},\frac{1}{\sqrt{2(n-1)}},\ldots, \frac{1}{\sqrt{2(n-1)}}\biggr)$, entropy becomes $H(\bm{p}) = -\biggl[\frac{1}{2} \ln\biggl(\frac{1}{2}\biggr) + \frac{1}{2} \ln\biggl(\frac{1}{2(n-1)}\biggr)\biggr] \approx \frac{1}{2} \ln n$ as $n \rightarrow \infty$. 
All influence is concentrated at the hub. On the other hand, for the regular graph, all nodes connected to all others and PEV as $\bm{u} = \left( \frac{1}{\sqrt{n}}, \frac{1}{\sqrt{n}}, \dots, \frac{1}{\sqrt{n}} \right)$ and thus entropy as $H = -n \left( \frac{1}{n} \ln \frac{1}{n} \right) = \ln n$. All nodes contribute equally. 
For scale-free network, PEV is highly concentrated on hubs, but some weight on peripheral nodes leads entropy as $ \frac{1}{2} \ln n < H < \ln n$ as $n \rightarrow \infty$ (Fig. \ref{IPR_Entropy_linear_dynamics}(d-f)). Note that for fix thresholds $\tau_{1}$, $\tau_{2}$ for IPR in Eq.~(\ref{threshold_scheme}) that split networks into delocalized, weakly localized, strongly localized (i.e., regions $r_{1}$, $r_{2}$, $r_{3}$), we numerically calibrate threshold for entropy measure $\tau_{1}$, $\tau_{2}$ so that the partitioning into regimes ($r_{1}$, $r_{2}$, $r_{3}$) mirrors the IPR-based partitioning, even though the entropy behaves oppositely.

\vspace{0.2cm}
\textbf{Experimental Set-up.} We trained a GCN model to predict the Shanon entropy values and corresponding states of localization and delocalization. As depicted in Fig. \ref {GCN_IPR_ENTROPY} \, (b, d), the model's three-class classification prediction accuracy and confusion metric show a drop in accuracy when identifying the weakly localized classes. However, the model's performance in the strongly localized and delocalized regimes is strong, resulting in an overall accuracy of $\sim 90\%$ in capturing these three states.

\vspace{0.2cm}
\textbf{Rényi Entropy.} 
It is essential to state that instead of Shanon entropy, one can consider Rényi entropy for the state identification as an alternative solution, which provides a generalized framework to quantify the uncertainty or spread of a probability distribution. The use of Rényi entropy to analyze localization properties in complex networks \cite{anand2009entropy} is a well-known problem. The Rényi entropy of order $q > 0$ ($q \neq 1$) is defined as:
\begin{equation}
H_q(\bm{p}) = \frac{1}{1 - q} \ln \left( \sum_{i=1}^{n} p_i^q \right) \, ,
\end{equation}
where $p_{i} = |u_{i}|^{2}$.

\vspace{0.2cm}
\textbf{Connection with Shannon Entropy.}
Rényi entropy \( H_q \) is a family of entropies parameterized by \( q \). Specifically, as the parameter $q \to 1$, the Rényi entropy converges to the Shannon entropy ($H$) as 
$$
\lim_{q \to 1} H_{q} = -\sum_{i = 1}^{n} p_{i} \ln p_{i} = H.
$$
For numerical stability when $q \approx 1$, we use $q = 1 \pm \epsilon$ with $\epsilon \ll 1$. Further, for $q = 2$, Rényi entropy takes the form
\[
H_{2} = \frac{1}{1-2} \ln \left( \sum_{i = 1}^n p_i^2 \right) = -\ln \left( \sum_{i = 1}^n p_i^2 \right)= -\ln (\text{IPR})
\] \,   
where $\text{IPR} = \sum_{i = 1}^n p_i^2$. Hence, the Rényi entropy of order \( q = 2 \) is logarithmically related to IPR values.

\end{document}